\UseRawInputEncoding
\documentclass[lettersize, journal]{IEEEtran}  

\usepackage{amssymb,amsfonts,amsmath,amscd}
\usepackage{cite}
\usepackage{graphicx} 
\usepackage{hyperref}
\usepackage{array}  
\usepackage{xcolor}  
\usepackage{subfigure}
\usepackage{acronym}
\usepackage{bm}
\usepackage{mleftright}
\usepackage{booktabs}
\usepackage{algpseudocode}
\usepackage{algorithmicx}
\usepackage{algorithm}
\usepackage{float}

\usepackage{textcomp}
\usepackage{stfloats}
\usepackage{url}
\usepackage{verbatim}
\usepackage{balance}

\hypersetup{
	colorlinks = true,
	citebordercolor=.,
	citecolor=green,
}

\def\BibTeX{{\rm B\kern-.05em{\sc i\kern-.025em b}\kern-.08em
		T\kern-.1667em\lower.7ex\hbox{E}\kern-.125emX}}

\algnewcommand\algorithmicinput{\textbf{Input:}}
\algnewcommand\INPUT{\item[\algorithmicinput]}
\algnewcommand\OUTPUT{\item[\textbf{Output:}]}

\newcolumntype{?}{!{\vrule width 0.2mm}}

\usepackage{tikz}
\usetikzlibrary{calc, arrows, shapes, positioning, decorations.pathreplacing,
	 calligraphy, shadings, angles, backgrounds, shapes.geometric, external}
\usetikzlibrary{shapes.geometric}
\tikzset{database/.style={cylinder,aspect=0.5,draw,rotate=90,path picture={
			\draw (path picture bounding box.160) to[out=180,in=180] (path picture bounding
			box.20);
			\draw (path picture bounding box.200) to[out=180,in=180] (path picture bounding
			box.340);
}}}

\usepackage{pgfplots}

\acrodef{IMU}{inertial measurement unit}
\acrodefplural{IMU}{inertial measurement units}
\acrodef{GP}{Gaussian process}
\acrodefplural{GP}{Gaussian processes}
\acrodef{SDE}{stochastic differential equation}
\acrodefplural{SDE}{stochastic differential equations}
\acrodef{LTI}{linear time-invariant}
\acrodef{SLAM}{simultaneous localization and mapping}
\acrodef{NEES}{normalized estimation error squared}

\makeatletter
\renewcommand*\env@matrix[1][*\c@MaxMatrixCols c]{%
	\hskip -\arraycolsep
	\let\@ifnextchar\new@ifnextchar
	\array{#1}}
\makeatother

\gdef\p{\boldsymbol{\Phi}}

\gdef\w{\boldsymbol{\varpi}}
\gdef\x{\boldsymbol{\xi}}
\gdef\e{\boldsymbol{\epsilon}}
\gdef\J{\bm{\mathcal{J}}}
\gdef\T{\bm{\mathcal{T}}}

\newcommand{\change}[1]{{\color{black} #1}}

\begin{document}
	
\title{Continuous-Time Radar-Inertial and Lidar-Inertial Odometry using a Gaussian Process Motion Prior}

\author{Keenan Burnett,~\IEEEmembership{Graduate Student Member,~IEEE}, Angela P. Schoellig,~\IEEEmembership{Member,~IEEE}, Timothy~D.~Barfoot,~\IEEEmembership{Fellow,~IEEE}
	\thanks{This work was supported in part by the Natural Sciences and Engineering Research Council of Canada (NSERC) and by an Ontario Research Fund: Research Excellence (ORF-RE) grant. \textit{(Corresponding author: Keenan Burnett)}}
	\thanks{Keenan Burnett and Timothy D. Barfoot are with the University of Toronto Institute for Aerospace Studies, Toronto, ON M3H5T6, Canada (e-mail: keenan.burnett@robotics.utias.utoronto.ca; tim.barfoot@utoronto.ca).}
	\thanks{Angela P. Schoellig is with the Technical University of Munich, 80333 Munich, Germany (e-mail: angela.schoellig@tum.de).}
}
	

\maketitle
\bibliographystyle{IEEEtran}

\begin{abstract}

In this work, we demonstrate continuous-time radar-inertial and lidar-inertial odometry using a Gaussian process motion prior. Using a sparse prior, we demonstrate improved computational complexity during preintegration and interpolation. We use a white-noise-on-acceleration motion prior and treat the gyroscope as a direct measurement of the state while preintegrating accelerometer measurements to form relative velocity factors. Our odometry is implemented using sliding-window batch trajectory estimation. To our knowledge, our work is the first to demonstrate radar-inertial odometry with a spinning mechanical radar using both gyroscope and accelerometer measurements. We improve the performance of our radar odometry by \change{43\%} by incorporating an IMU. Our approach is efficient and we demonstrate real-time performance. Code for this paper can be found at: \url{https://github.com/utiasASRL/steam_icp}

\end{abstract}

\begin{IEEEkeywords}
	Localization, Range Sensing, Mapping, Continuous-Time
\end{IEEEkeywords}

\section{Introduction}


\IEEEPARstart{I}{n} large-scale outdoor mapping and localization, focus has shifted towards improving robustness and reliability under challenging conditions such as sparse or degenerate geometry, aggressive motion, and adverse weather. Radar is a promising alternative to lidar as its longer wavelength enables it to be robust to dust, fog, rain, and snow. Although it has been shown in prior work that lidar localization can still function under moderate levels of precipitation \cite{burnett_ral22}, it is possible that radar will perform better than lidar under more extreme weather conditions. Furthermore, radar-based localization may still prove valuable as a redundant backup system in safety-critical applications. Our goal in this work is to tackle the problem of aggressive motion using continuous-time state estimation and an inertial measurement unit (IMU). In addition, we apply our approach to radar-inertial odometry so as to address adverse weather conditions. Our secondary goal is to reduce the performance gap between radar and lidar odometry through the incorporation of an IMU.

\acused{IMU}

Inertial measurement units play an important role in many robotic estimation systems and are often fused with low-rate exteroceptive measurements from sensors such as a camera. The addition of an \ac{IMU} encourages the estimated trajectory to be locally smooth and helps the overall pipeline to be more robust to temporary failures of the exteroceptive measurements. Ordinarily, in discrete-time batch state estimation, one would need to estimate the state at each measurement time. However, due the high rate of  \acp{IMU}, the number of measurement times can become quite large and consequently the computational requirements can become too expensive for real-time operation. 





\begin{figure}[t]
	\centering
	\includegraphics[trim=0 0 0 0, clip, width=1.0\columnwidth]{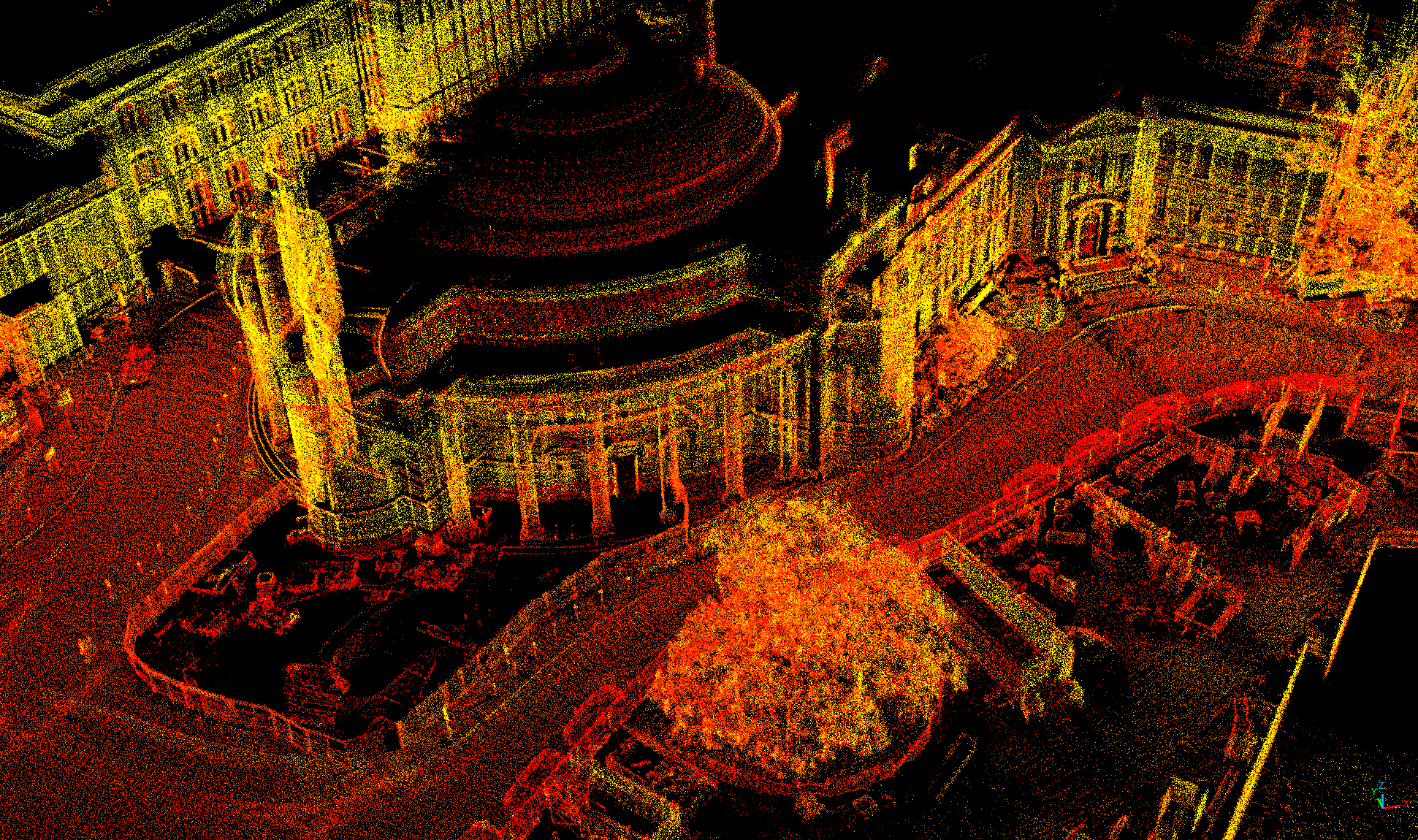}
	\caption{A lidar map generated of the University of Toronto obtained during a sequence of the Boreas dataset. This high-quality map is generated as a byproduct of our odometry pipeline. The pointcloud is colored by intensity.}
	\label{fig:utias_front}
\end{figure}

In order to address this problem, Lupton and Sukkarieh \cite{lupton_tro12} proposed to preintegrate \ac{IMU} measurements between pairs of consecutive camera measurements in order to combine them into a single relative motion factor. This significantly improves runtime since we now only need to estimate the state at each camera measurement. Forster et al. \cite{forster_tro17} then showed how to perform on-manifold preintegration within the space of 3D rotations, $SO(3)$. Recently, Brossard et al. \cite{brossard_tro22} demonstrated how to perform on-manifold preintegration within the space of extended poses, $SE_2(3)$, which captures the uncertainty resulting from IMU measurements more consistently.

Classical preintegration was designed to address the problem of combining a low-rate sensor with a high-rate inertial sensor. However, in some cases we must work with multiple high-rate sensors such as a lidar or radar and an IMU. While lidar sensors typically spin at around 10Hz, the laser measurements are acquired at a much higher rate, on the order of 10kHz. At this rotational rate (10Hz), the motion of the robot causes the pointclouds to become distorted due to the scanning-while-moving nature of the sensor. In our previous work, we addressed this motion-distortion effect using continuous-time point-to-point factors and a Gaussian process motion prior \cite{burnett_ral22}. This allowed us to estimate the pose and body-centric velocity of the sensor while simultaneously undistorting the data. Given this previously demonstrated success, we are motivated to apply our continuous-time techniques to radar-inertial and lidar-inertial odometry. \change{It is possible to employ a constant-velocity assumption \cite{vizzo_ral23} when the motion of a robot is relatively smooth, such as in the case of heavy ground robots. However, when working with highly dynamic motions, such as in the case of drones or walking robots, continuous-time approaches present considerable advantages. We will demonstrate this in the experimental results section using the Newer College Dataset.}

Our aim is to treat the lidar, radar, and IMU all as high-rate measurements using continuous-time estimation. We are thus faced with the choice of picking a suitable Gaussian process motion prior. In this work, we choose to use a white-noise-on-acceleration motion prior \cite{anderson_iros15}. In this setup, angular velocity and body-centric linear velocity are a part of our state. As such, we treat the gyroscope as a direct measurement of the state rather than preintegrating it. However, since acceleration is not a part of the state, we still need to preintegrate the accelerometer measurements to form relative velocity factors. We only need to integrate the accelerometer measurements once as we rely on the Gaussian process estimation framework to do the remaining integration into position. Other motion prior factors are also possible such as white-noise-on-jerk \cite{tang_ral19} or the Singer prior \cite{wong_ral20b}, both of which include body-centric acceleration in the state. In our experiments, we observed that including these higher-order derivatives in the state sometimes improved performance. However, the effect was not consistent across datasets. Furthermore, we observed that including acceleration in the state made the overall pipeline less reliable and thus we opted to not include it.


Another work that employed Gaussian processes in continuous-time state estimation is that of Le Gentil and Vidal-Calleja \cite{legentil_ijrr23}. They proposed to model the state using six independent Gaussian processes, three for angular velocity, and three for linear acceleration. They estimate the state of the Gaussian processes at several inducing points given the measurements and then analytically integrate these Gaussian processes to form relative motion factors on position, velocity, and rotation in a manner similar to classical preintegration. They use a squared exponential kernel, which misses out on the potential benefits of using a sparse kernel. As a result, the computational complexity of their approach is $O(J^3 + J^2N)$ while ours is only $O(J + N)$ where $J$ is the number of inducing points and $N$ is the number of query times. In our approach, all six degrees of freedom are coupled through the motion prior. Consequently, our approach has the potential to provide better calibrated covariance estimates. Their exponential kernel results in a fully-connected factor graph, and so dividing a longer trajectory into a sequence of local chunks effectively drops connections from the graph. In our approach, dividing a longer trajectory into a sequence of local GPs is less of an approximation due to the Markovian nature of the state that results from a sparse kernel. In addition, our approach can still perform continuous-time lidar odometry during a period of IMU measurement dropout by falling back on the Gaussian process motion prior. We consider our approach to be tightly coupled since we include IMU measurements in the pointcloud alignment optimization directly whereas their approach uses the IMU to first undistort the scans prior to the alignment optimization. Another important difference is how we compensate for motion distortion. \cite{legentil_ijrr23} undistorts pointclouds using the upsampled preintegrated IMU measurements whereas our approach uses the posterior of our continuous-time trajectory, which includes both IMU and lidar measurements. In Figure~\ref{fig:utias_front}, we provide an example of a map generated using our approach.

Finally, it should be noted that our work focuses on the back-end continuous-time state estimation and is compatible with other works that focus on the front-end pointcloud preprocessing, submap keyframing strategy, and efficient map storage improvements \cite{xu_tro22, chen_icra23}. Also, our framework supports adding additional continuous-time measurement factors such as wheel odometry and Doppler velocity measurements \cite{wu_ral22}. We provide experimental results in the autonomous driving domain and using a hand-held sensor mast demonstrating that our approach is generalizable to different domains. To summarize, our contributions are as follows:
\begin{itemize}
	\item We demonstrate continuous-time lidar-inertial and radar-inertial odometry using a Gaussian process motion prior where the preintegration cost is linear in the number of estimation times.
	\item We provide experimental results of our real-time approach on three datasets: KITTI-raw \cite{geiger_ijrr13}, Boreas \cite{burnett_ijrr23} and the Newer College Dataset \cite{ramezani_iros20}.
	\item We demonstrate radar-inertial odometry with a spinning mechanical radar using both gyroscope and accelerometer measurements. To our knowledge, this has not been previously demonstrated in the literature.
	\item We provide a detailed comparison of the performance of lidar-inertial and radar-inertial odometry across varying seasonal and weather conditions.
\end{itemize}

\section{Related Work}

For a detailed literature review on lidar odometry and lidar-inertial odometry, we refer readers to the recent survey by Lee et al. \cite{lee_arxiv23}.


\subsection{Lidar Odometry}

Lidar odometry methods can be broadly classified into feature-based approaches, which seek to extract and match sparse geometric features, and direct methods, which work directly with raw lidar pointclouds. Direct methods usually rely on a variation of iterative closest point (ICP) to match pairs of pointclouds \cite{pomerleau_ar13}. Due to the large number of points produced by modern lidar sensors ($\sim$100k points per scan), these methods incur a heavy computational load. In order to enable real-time operation, recent methods rely on coarse voxelization and efficient data structures for map storage and retrieval \cite{dellenbach_icra22, vizzo_ral23, xu_tro22, chen_icra23}. Care must also be taken to tune ICP parameters such as the maximum point-to-point matching distance to ensure reliable and fast convergence.  There are also several ICP variants to choose from such as point-to-point, point-to-plane, and Generalized ICP \cite{pomerleau_ar13, segal_rss09}. Our work can be considered a direct method since we do not make use of feature extraction and matching.

Examples of feature-based methods include LOAM \cite{zhang_ar17}, which matches edge and plane features, and SuMa++ \cite{chen_iros19}, which matches surfels using ICP aided by semantic segmentation labels from a neural network. Feature-based methods tend to work well in structured environments but may experience a drop in performance in unstructured environments.

\subsection{Lidar-Inertial Odometry}


Prior works have leveraged inertial measurement units (IMUs) to address several shortcomings of lidar-only odometry. Firstly, IMU measurements can be used to compensate for the motion-distortion effect of lidar sensors \cite{zhang_ar17}. Furthermore, IMUs can enable lidar-inertial odometry to tackle trajectories with more aggressive motion and potentially degenerate geometry. Previous lidar-inertial odometry literature can then be sorted based on the degree of integration between the lidar and IMU modalities ranging from loosely coupled to tightly coupled. Tightly coupled methods incorporate IMU data into the pointcloud alignment optimization directly.

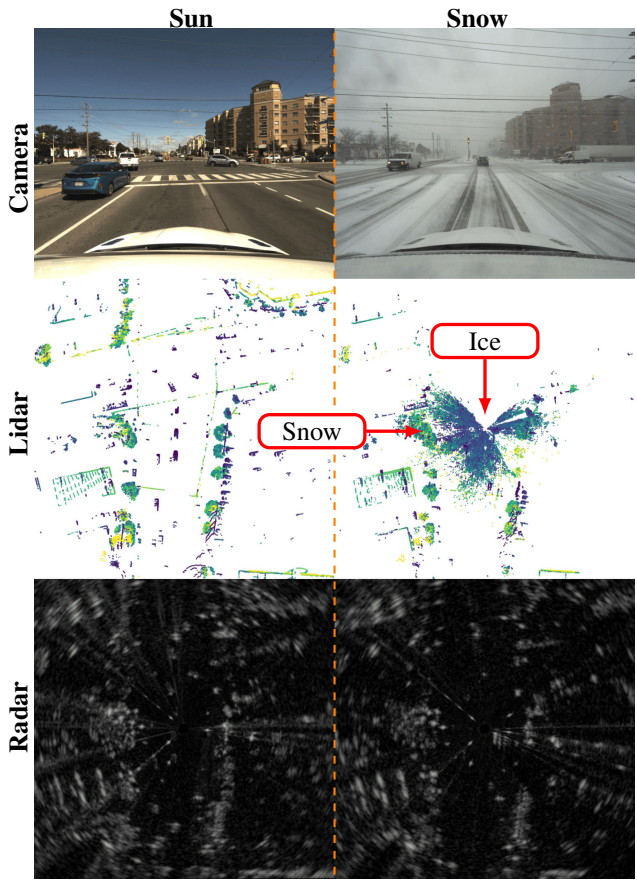
\begin{figure} [t]
	\centering
	\begin{tikzpicture}[arrow/.style={>=latex,red, line width=1.25pt}, block/.style={rectangle, draw,
						minimum width=4em, text centered, rounded corners, minimum height=1.25em, line width=1.25pt, inner sep=2.5pt}]
		
%
%
	\node[inner sep=0pt] (cam1)
	{\includegraphics[width=0.9\columnwidth]{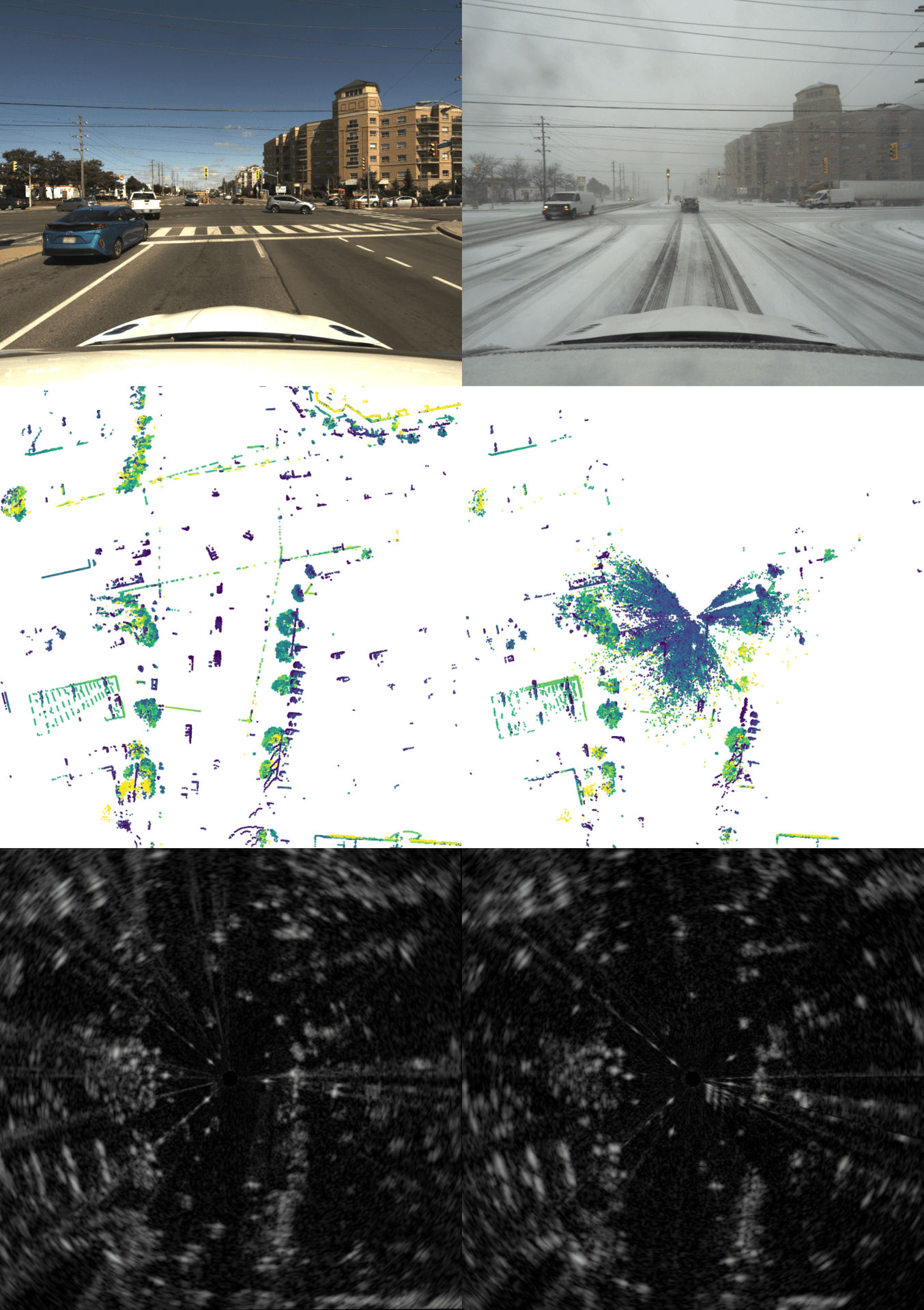}};
	
	\node (text1) [left of=cam1, xshift=-32mm, yshift=38mm, rotate=90] {\textbf{Camera}};
	\node (text2) [left of=cam1, xshift=-32mm, yshift=4mm, rotate=90] {\textbf{Lidar}};
	\node (text3) [left of=cam1, xshift=-32mm, yshift=-35mm, rotate=90] {\textbf{Radar}};
	\node (text4) [above of=cam1, yshift=48mm, xshift=-19mm] {\textbf{Sun}};
	\node (text5) [above of=cam1, yshift=48mm, xshift=19mm] {\textbf{Snow}};
	
	\node [block,  draw=red, fill=white, left of=cam1, xshift=30mm, yshift=15mm] (ice) {Ice};
	\def \L{0.75};
	\coordinate (r3) at ($ (ice.south) + (0, -0.75) $);
	\draw[->, arrow] (ice.south) -- (r3) {};
	
	\node [block,  draw=red, fill=white, left of=cam1, xshift=7mm, yshift=3mm] (snow) {Snow};
	\coordinate (r1) at ($ (snow.east) + (\L, 0) $);
	\draw[->, arrow] (snow.east) -- (r1) {};
	\coordinate (l1) at ($ (cam1) + (0, +5.6) $);
	\coordinate (l2) at ($ (cam1) + (0, +.6) $);
	\coordinate (l3) at ($ (cam1) + (0, -5.6) $);
	\coordinate (l4) at ($ (cam1) + (0, -.0) $);
	\draw[-, orange, dashed, thick] (l1) -- (l2);
	\draw[-, orange, dashed, thick] (l3) -- (l4);

	\end{tikzpicture}
	\caption{This figure illustrates the differences in camera, lidar, and radar data during a sunny day and during a snowstorm. The lidar data here is colored by elevation. During snowfall, the lidar data becomes corrupted with noise and a section becomes blocked by ice but the radar data appears unaffected.}
	\label{fig:weather}
\end{figure}

LOAM \cite{zhang_ar17} is an example of a loosely coupled approach where an IMU is used to undistort lidar pointclouds for use in ICP within a discrete-time state estimation framework where the IMU preintegration may be used as an initial guess for ICP. LIO-SAM \cite{shan_iros20} and LION \cite{tagliabue_er21} are examples of loosely coupled approaches that undistort lidar data using an IMU and then include both relative pose estimates from ICP and preintegrated IMU measurements in a factor graph. DLIO \cite{chen_icra23} is a recent example of a loosely coupled approach where an IMU is used to undistort lidar data. Preintegrated IMU measurements are then subsequently combined with pose estimates from ICP using a hierarchical geometric observer. LIOM \cite{ye_icra19} is an example of a tightly coupled lidar-inertial odometry using a factor graph. FAST-LIO2 \cite{xu_tro22} is another tightly coupled approach that uses an iterated extended Kalman filter. In our tightly coupled approach, we combine continuous-time point-to-plane factors with direct gyroscope factors and preintegrated velocity factors using a fixed-lag smoother and a Gaussian process motion prior.

Even after incorporating an IMU, some challenges remain such as handling harsh enviromental conditions such as dust, fog, rain, and snow that can adversely affect lidar data. Figure~\ref{fig:weather} depicts an example where lidar data is affected by snow and ice build-up. In order to tackle these problems, radar is being investigated as a potential alternative to lidar.

\subsection{Radar Odometry}

Radar mapping and localization is a long-standing research area in robotics. The first methods to demonstrate radar-based localization relied on reflective beacons installed in the environment in order to boost the signal-to-noise ratio between the detected target power and background clutter power \cite{clark_icra98, dissanayake_tro01}. Due to the noise inherent to radar, it is challenging to directly apply methods designed for lidar. 

Several methods have been proposed to deal with the high amount of noise in radar measurements. Jose and Adams \cite{jose_iros04} proposed to estimate the probability of target presence and to include radar cross section in their \ac{SLAM} setup. Checchin et al. \cite{checchin_fsr10} proposed to densely match radar scans using the Fourier-Mellin transform, a correlative scan-matching approach. Vivet et al. \cite{vivet_sens13} and Kellner et al. \cite{kellner_itsc13} proposed to use the relative Doppler velocity measurements to estimate the instantaneous egomotion. Callmer et al. \cite{callmer_eurasip11} proposed to leverage features originally designed for vision to enable landmark-based \ac{SLAM}. Schuster et al. \cite{schuster_itsc16} subsequently refined this approach by designing bespoke radar feature descriptors. Cen and Newman demonstrated accurate large-scale radar odometry using a spinning mechanical radar \cite{cen_icra18}. Hong et al. \cite{hong_iros20} demonstrated large-scale radar SLAM in all weather conditions. Recently, there has been a resurgence of research into improving and refining radar-based localization. Currently, the state of the art for radar odometry with a spinning mechanical radar is CFEAR, which extracts only the $k$ strongest detections on each scanned azimuth and subsequently matches the live radar scan to a sliding window of keyframes in a manner similar to ICP \cite{adolfsson_tro22}. For a more detailed review of radar-based localization, we refer readers to the survey by Harlow et al. \cite{harlow_arxiv23}.

Recent works have benefited from advancements in radar sensors where frequency modulated continuous wave (FMCW) radar sensors now possess improved range resolution, azimuth resolution, and some even provide target elevation in 3D. Furthermore, many FMCW radar sensors also provide a relative Doppler velocity measurement for each target. However, these new radar sensors still produce sparse and noisy pointclouds. As such, there has been a concerted effort to develop and improve radar-inertial odometry to address these challenges.

\subsection{Radar-Inertial Odometry}

Radar-inertial odometry literature tends to focus on either low-cost consumer-grade radar for resource-constrained applications such as UAVs or automotive-grade radar for large-scale outdoor applications. Examples of prior work using consumer-grade radar include \cite{kramer_icra20, almalioglu_sens20, park_ral21, doer_iros21, michalczyk_iros22, michalczyk_icra23, chen_ral23, huang_arxiv23}. Examples of prior work using automotive radar include \cite{ng_iros21, zhuang_ral23, kubelka_arxiv23}. The work that is closest to ours is that of Ng et al. \cite{ng_iros21}, which demonstrated continuous-time radar-inertial odometry using four automotive radars. In our approach, we use a Gaussian process motion prior to enable continuous-time trajectory estimation whereas their approach uses cubic B-splines. Another important difference is that our approach uses continuous-time point-to-plane factors whereas their approach uses only the Doppler velocity measurements from each sensor in conjunction with an IMU. The radar that we use in this work does not currently support Doppler velocity measurements. However, we have previously shown that our continuous-time motion prior supports incorporating these measurement factors when they are available \cite{wu_ral22}. To our knowledge, our work is the first to demonstrate  radar-inertial odometry using a spinning mechanical radar. Previous works have used a combination of wheel odometry and single-axis gyroscopes \cite{rouveure_radar09, mullane_icra12, vivet_ijars13} to compensate for motion distortion and as a prediction step in a filter while our work fuses both gyroscope and accelerometer measurements with point-to-point radar factors using our continuous-time framework.

\begin{figure}[t]
	\centering
	\begin{tikzpicture}[
		line cap=round,
		graphedge/.style={>=latex, line width=1pt},
		state/.style={draw, thick, line width=1pt, isosceles triangle,isosceles triangle apex angle=45, minimum size=4mm, inner sep=0pt, outer sep=0pt, fill=white},
		interpolated/.style={state, minimum size=2mm},
		dot/.style={draw, inner sep=0, circle, fill, minimum size=2mm, line width=0pt},
		mappoint/.style={dot, black, minimum size=1mm, outer sep=1mm},
		vfactor/.style={dot, green},
		preintfactor/.style={dot, magenta},
		pfactor/.style={dot, red},
		icpfactor/.style={dot, blue},
		every label/.style={align=left}
		]
		
		\def \mx{0.80};
		
		\draw[graphedge, ->] (0, 0) -- (\mx*9, 0) node[right]{$t$};
		\foreach \i in {0, 1, ..., 8} \draw[graphedge] (\mx*\i,0) -- (\mx*\i, .1);
		
		\node[label=left:{Radar}] (t1) at (0, 1) {};
		\node[label=left:{IMU}] (t2) at (0, 2) {};
		\node[label=left:{State}] (t3) at (0, 3) {};

		\foreach \i in {0, 1, ..., 32} \draw[graphedge, blue] (\mx*\i/4,1) -- (\mx*\i/4, 1.1);
		\foreach \i in {0, 1, ..., 16} \draw[graphedge, green] (\mx*\i/2 + 0.125 ,2) -- (\mx*\i/2 + 0.125 , 2.1);
		\foreach \i in {0, 1, ..., 2} \node[interpolated] at (\mx*\i*4,3) {};
		
	\end{tikzpicture}
	\caption{This figure illustrates our asynchronous sensor timing where states are estimated for each scan obtained by the radar. Our radar outputs measurements at 1600Hz while our lidar outputs unique timestamps at roughly 40kHz and our IMU outputs measurements at 200Hz.}
	\label{fig:sensor_timing_diagram}
\end{figure}
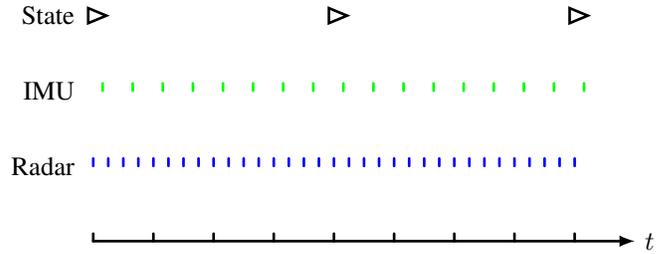

The radar and lidar sensors used in this work both rely on mechanical actuation to cover the entire field of view around the robot. As a consequence, these sensors suffer from a motion-distortion effect while moving. In addition, it is difficult to combine these sensors with asynchronous IMU measurements without resorting to ad-hoc interpolation schemes. These challenges motivate the use of continuous-time trajectory estimation. Figure~\ref{fig:sensor_timing_diagram} illustrates the high-rate and asynchronous nature of the sensor measurements.







\subsection{Continuous-Time State Estimation}

There are two main classes of continuous-time approaches, parametric approaches relying on temporal basis functions and non-parametric approaches such as Gaussian processes. Two popular examples of parametric approaches are linear interpolation and cubic B-splines.

Linear interpolation is often performed in the Lie algebra between pairs of discrete samples of the trajectory. This approach assumes a constant velocity between pairs of poses and relies on sampling the trajectory at a sufficiently high rate to support dynamic motions. These approaches sometimes upsample the estimated trajectory using splines to remove motion distortion from pointclouds. Smoothness factors may be included to penalize acceleration between pairs of poses. Examples of linear interpolation approaches include \cite{bosse_tro12, nguyen_ral23, zheng_arxiv23}. CT-ICP \cite{deschaud_icra18} is another example of linear interpolation where their innovation was to parametrize each lidar scan as a pair of poses at the start and end of the scan and to model the motion during a scan with constant velocity while allowing trajectory discontinuities between scans.

Parametric approaches seek to represent the trajectory using a finite set of temporal basis functions. Previous examples of parametric approaches include \cite{furgale_ijrr15, patron_ijcv15}. Recent examples of parametric approaches applied to lidar odometry and lidar-inertial odometry include \cite{droeschel_icra18, quenzel_iros21, park_tro21, lv_tmech23, lang_ral23} all of which use B-splines.

Non-parametric approaches such as Gaussian processes (GPs) seek to model a continuous-time trajectory implicitly given a set of measurements of the state. The state at a set of estimation times can then be determined by performing Gaussian process regression. These estimation times may be chosen independently of the measurement times. The posterior GP may then be queried at any time of interest. In prior work, it was shown that for vector spaces, a linear time-varying stochastic differential equation can be interpreted as a Gaussian process and batch continuous-time trajectory estimation can be performed efficiently thanks to the exact sparsity of the inverse kernel matrix owing to the Markovian nature of the state \cite{anderson_ar15}. Instead of the usual cubic time complexity for Gaussian process regression, this approach enables linear time complexity. Furthermore, this work showed that posterior interpolation could be performed as an O(1) operation. Subsequently, Anderson and Barfoot extended this approach to work with $SE(3)$ where the trajectory is divided into a sequence of local GPs \cite{anderson_iros15}. Recently, Le Gentil and Vidal-Calleja employed Gaussian processes in order to model the linear acceleration and angular velocities in continuous-time given a set of IMU measurements \cite{legentil_tro20, legentil_ijrr23}. They then used their estimated GP to upsample IMU measurements towards undistorting pointclouds and to provide improved preintegration measurements for inertial-aided state estimation. One appealing aspect of our approach is that we start from a physically motivated prior: white noise on acceleration or constant velocity. Furthermore, compared to the linear interpolation approaches, the Gaussian process prior provides a principled manner to construct motion priors and perform interpolation. In addition, the hyper-parameters of the GP can be learned from a training set using maximum likelihood, enabling a data-driven approach to fine-tune the GP to each application. Determining the spacing of control points is an important engineering challenge in the use of B-splines which can be avoided by using Gaussian processes instead. \change{For a comparison of splines and Gaussian processes, we refer to Johnson et al. \cite{johnson_arxiv24}. For a survey on continuous-time state estimation, we refer to Talbot et al. \cite{talbot_arxiv24}.}

\section{Continuous-Time Trajectory Estimation}


In this section, we review relevant prior work \cite{anderson_ar15, anderson_iros15} on continuous-time trajectory estimation using Gaussian processes. We begin with the following nonlinear time-varying stochastic differential equation,
\begin{subequations}\label{eq:wnoa}
	\begin{align}
		\mathbf{\dot{T}}(t) &= \boldsymbol{\varpi}(t)^\wedge \mathbf{T}(t)  \\
		\dot{\boldsymbol{\varpi}}(t) &= \mathbf{w}^\prime(t), \quad \mathbf{w}^\prime(t) \sim \mathcal{GP}(\mathbf{0}, \bm{Q}^\prime \delta (t - t^\prime))
	\end{align}
\end{subequations} where $\mathbf{T}(t) \in SE(3)$ is the pose, $\w(t) = [\boldsymbol{\nu}^T~\boldsymbol{\omega}^T]^T \in \mathbb{R}^6$ is the body-centric velocity consisting of a linear $\boldsymbol{\nu}(t)$ and angular $\boldsymbol{\omega}(t)$ component, and $\mathbf{w}^\prime(t)$ is a white-noise Gaussian process where $\bm{Q}^\prime$ is the symmetric positive-definite power-spectral density matrix. We refer to this as white-noise-on-acceleration due to white noise being injected on the body-centric acceleration $\dot{\boldsymbol{\varpi}}(t)$. The above nonlinear time-varying stochastic differential equation is then approximated using a sequence of local linear time-invariant stochastic differential equations \cite{anderson_iros15}. Between pairs of estimation times, $t_k$ and $t_{k+1}, k = 0 \ldots K-1$, local pose variables are defined in the Lie algebra $\boldsymbol{\xi}_k(t) \in \mathfrak{se}(3)$ such that 

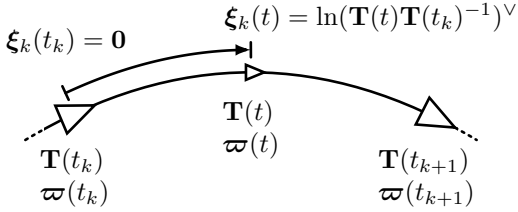
\begin{figure}[t]
	\centering
	\begin{tikzpicture}[
		line cap=round,
		graphedge/.style={>=latex, line width=1pt},
		state/.style={draw, thick, line width=1pt, isosceles triangle,isosceles triangle apex angle=45, minimum size=4mm, inner sep=0pt, outer sep=0pt, fill=white},
		interpolated/.style={state, minimum size=2mm},
		dot/.style={draw, inner sep=0, circle, fill, minimum size=2mm, line width=0pt},
		mappoint/.style={dot, black, minimum size=1mm, outer sep=1mm},
		vfactor/.style={dot, green},
		preintfactor/.style={dot, magenta},
		pfactor/.style={dot, red},
		icpfactor/.style={dot, blue},
		every label/.style={align=left}
		]
		
 		\begin{scope}
			\clip (0.25,0) rectangle (5.75, 2);
			\draw[graphedge] (0, 0) .. controls (2, 1.2) and (4, 1.2) .. (6, 0)  node (a1) [pos=0.1,state, rotate=23]{} node (a2) [pos=0.5,interpolated]{} node (a3) [pos=0.9,state, rotate=-27]{};
		\end{scope}
		
		\begin{scope}
			\clip (-0.5,-0.5) rectangle (0.25, 2);
			\draw[graphedge, dotted] (0, 0) .. controls (2, 1.2) and (4, 1.2) .. (6, 0);
		\end{scope}

		\begin{scope}
			\clip (5.75,-0.5) rectangle (6.5, 2);
			\draw[graphedge, dotted] (0, 0) .. controls (2, 1.2) and (4, 1.2) .. (6, 0);
		\end{scope}
		
		\node[] (t1) [below=of a1.south, yshift=-2mm, label={$\mathbf{T}(t_k)$ \\ $\w(t_k)$}] {};
		\node[] (t2) [below=of a2.south, yshift=-2mm, label={$\mathbf{T}(t)$ \\ $\w(t)$}] {};
		\node[] (t3) [below=of a3.south, yshift=-2mm, label={$\mathbf{T}(t_{k+1})$ \\ $\w(t_{k+1})$}] {};
		
 		\path (0, 0) .. controls (2, 1.2) and (4, 1.2) .. (6, 0) {\foreach \i in {1,...,40} {  coordinate[pos=0.10+0.40*\i/40] (p\i) } } ;
			
	  \draw[graphedge, ->] ($(p1)!.3cm!90:(p2)$) 
		{ \foreach \i [count=\j from 3] in {2,...,39} {-- ($(p\i)!.3cm!90:(p\j)$) } }
		-- ($(p40)!.3cm!-90:(p39)$);
		
		\draw[graphedge] ($(p1)!.4cm!90:(p2)$) -- ($(p1)!.2cm!90:(p2)$);
		\draw[graphedge] ($(p40)!.4cm!-90:(p39)$) -- ($(p40)!.2cm!-90:(p39)$);
		
		\node (x1) [above=of a1.north, yshift=-7mm, label={$\boldsymbol{\xi}_k(t_k) = \mathbf{0}$}] {};
		\node (x2) [above=of a2.north, yshift=-9mm, xshift=16mm, label={$\boldsymbol{\xi}_k(t) = \ln(\mathbf{T}(t) \mathbf{T}(t_k)^{-1})^\vee$}] {};
		
	\end{tikzpicture}
	\caption{This figure illustrates the local variable $\x_k(t)$ which is defined in the tangent space of the pose at time $t_k$. The larger triangles denote the estimated states while the smaller triangle denotes the interpolated state at time $t$.}
	\label{fig:steam_factor_graph}
\end{figure}

\begin{equation}
	\mathbf{T}(t) = \exp(\boldsymbol{\xi}_k(t)^\wedge) \mathbf{T}(t_k). \label{eq:loc_var}
\end{equation} The local kinematic equations are then defined as
\begin{equation} \label{eq:local_lti}
	\ddot{\boldsymbol{\xi}}_k(t) = \mathbf{w}_k(t), \quad \mathbf{w}_k(t) \sim \mathcal{GP}(\mathbf{0}, \bm{Q} \delta (t - t^\prime)).
\end{equation} This approximation of \eqref{eq:wnoa} holds so long as the process noise is small and the rotational motion between pairs of estimation times is also small. The local Markovian state variables are defined as
\begin{equation} \label{eq:local_markov}
	\boldsymbol{\gamma}_k(t) = \begin{bmatrix}
		\boldsymbol{\xi}_k(t) \\ \dot{\boldsymbol{\xi}}_k(t)
	\end{bmatrix}.
\end{equation} The local LTI SDE defined by \eqref{eq:local_lti}, \eqref{eq:local_markov} is then stochastically integrated to arrive at the following local GP:
\begin{align}
	\boldsymbol{\gamma}_k(t) \sim \mathcal{GP}(&\boldsymbol{\Phi}(t, t_k) \check{\boldsymbol{\gamma}}_k(t_k)), \\ &\boldsymbol{\Phi}(t, t_k) \check{\mathbf{P}}(t_k) \boldsymbol{\Phi}(t, t_k)^T +  \mathbf{Q}_k   ), \nonumber
\end{align} where
\begin{equation}
	\boldsymbol{\Phi}(t, t_k) = \begin{bmatrix} \mathbf{1} & (t - t_k) \mathbf{1} \\ \mathbf{0} & \mathbf{1} \end{bmatrix}
\end{equation} is the transition function,
\begin{equation}
	\mathbf{Q}_k = \begin{bmatrix} \frac{1}{3} (t - t_k)^3 \bm{Q} & \frac{1}{2} (t - t_k)^2 \bm{Q} \\ \frac{1}{2}(t - t_k)^2 \bm{Q} & (t - t_k) \bm{Q} \end{bmatrix}
\end{equation} is the covariance between two times, $t$, $t_k$, and $\check{\boldsymbol{\gamma}}_k(t_k)$, $\check{\mathbf{P}}(t_k)$ are the initial mean and covariance at $t = t_k$, the starting point of the local variable. Over a sequence of estimation times, $t_0 < t_1 < \cdots < t_K$, the kernel matrix can be written as
\begin{equation}
	\check{\mathbf{P}} = \text{cov}(\delta \mathbf{x}) = \mathbf{A} \mathbf{Q} \mathbf{A}^T,
\end{equation} where
\begin{equation} \label{eq:lifted_trans_inv2}
	\mathbf{A}^{-1} = \begin{bmatrix} 
		\mathbf{1} \\ 
		-\p(t_1,t_0)  & \ddots \\
		& \ddots &  \mathbf{1} \\
		& & -\p(t_K,t_{K-1}) & \mathbf{1}
	\end{bmatrix}
\end{equation} is the inverse of the lifted transition matrix and $\mathbf{Q} = \text{diag}(\check{\mathbf{P}}_0, \mathbf{Q}_1, \cdots, \mathbf{Q}_K)$. Even though the kernel matrix is dense, the inverse kernel matrix $\check{\mathbf{P}}^{-1} = \mathbf{A}^{-T} \mathbf{Q}^{-1} \mathbf{A}^{-1}$ is block-tridiagonal. The exact sparsity of the inverse kernel matrix is what allows us to perform batch trajectory estimation as exactly sparse Gaussian process regression. As a result, the computation for batch trajectory estimation scales linearly with the number of estimation times. 

In order to convert our continuous-time formulation into a factor graph, we construct a sequence of motion prior factors between pairs of estimation times,
\begin{subequations}
\begin{align}
	J_{v,k} &= \frac{1}{2} \mathbf{e}_{v,k}^T \mathbf{Q}_k^{-1} \mathbf{e}_{v,k}, \label{eq:prior1} \\
	\mathbf{e}_{v,k} &= \boldsymbol{\gamma}_k(t_{k+1}) - \check{\boldsymbol{\gamma}}_k(t_{k+1}) \nonumber \\
	&- \boldsymbol{\Phi}(t_{k+1}, t_k) (\boldsymbol{\gamma}_k(t_k) - \check{\boldsymbol{\gamma}}_k(t_k)), \label{eq:prior2}
\end{align}
\end{subequations} where \change{$J$ denotes a cost factor, $\mathbf{e}$ denotes an error function,} and $\check{\boldsymbol{\gamma}}_k(t) = E [ \boldsymbol{\gamma}_k(t) ]$ is the prior mean. In the absence of exogenous control inputs, $\check{\boldsymbol{\gamma}}_k(t) = \boldsymbol{\Phi}(t, t_k) \check{\boldsymbol{\gamma}}_k(t_k)$ and so \eqref{eq:prior2} simplifies to
\begin{equation}
		\mathbf{e}_{v,k} = \boldsymbol{\gamma}_k(t_{k+1}) - \boldsymbol{\Phi}(t_{k+1}, t_k) \boldsymbol{\gamma}_k(t_k). \label{eq:prior3}
\end{equation}

Now, to translate this prior factor, which is defined in terms of the local variables, into the global variables, we first rearrange \eqref{eq:loc_var} as
\begin{equation}
	\boldsymbol{\xi}_k(t) = \ln(\mathbf{T}(t) \mathbf{T}(t_k)^{-1})^\vee. \label{eq:local_convert_pose}
\end{equation} We then convert from body-centric velocity using:
\begin{equation}
	\dot{\boldsymbol{\xi}}_k(t) = \J(\boldsymbol{\xi}_k(t))^{-1} \boldsymbol{\varpi}(t), \label{eq:local_convert_vel}
\end{equation} where $\J$ is the left-hand Jacobian of $SE(3)$. From \eqref{eq:local_convert_pose}, \eqref{eq:local_convert_vel}, we can then define the local Markovian variable in terms of the global variables with

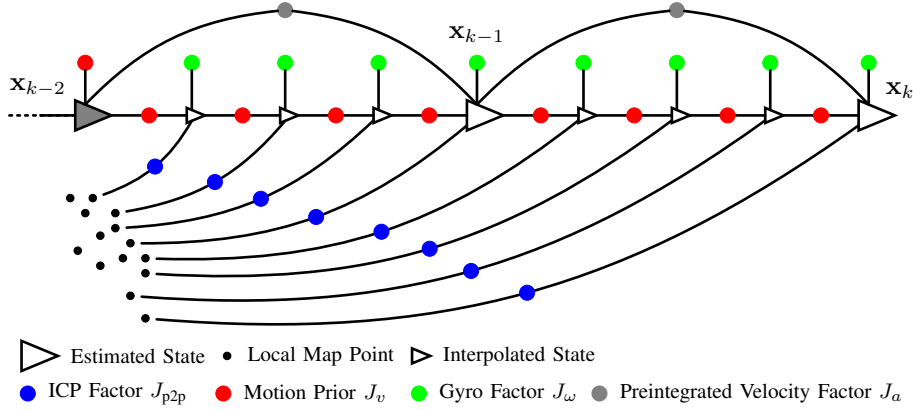
\begin{figure*}[t]
	\centering
	\begin{tikzpicture}[
		line cap=round,
		graphedge/.style={>=latex, line width=1pt},
		state/.style={draw, thick, line width=1pt, isosceles triangle,isosceles triangle apex angle=45, minimum size=4mm, inner sep=0pt, outer sep=0pt, fill=white},
		interpolated/.style={state, minimum size=2mm},
		dot/.style={draw, inner sep=0, circle, fill, minimum size=2mm, line width=0pt},
		mappoint/.style={dot, black, minimum size=1mm, outer sep=1mm},
		vfactor/.style={dot, green},
		preintfactor/.style={dot, gray},
		pfactor/.style={dot, red},
		icpfactor/.style={dot, blue},
		every label/.style={align=left}
		]
		
		\node[] (p1) {};
		\node[] (p2) [right= of p1.center] {};
		\draw[graphedge, dotted] (p1) -- (p2);
		\begin{scope}
			\clip (p2) circle (6mm);
			\draw[graphedge] (p1) -- (p2);
		\end{scope}
		
		\def \dx{0mm};
		\def \vy{0.7};
		\node[state, fill=gray, label=above left:{$\mathbf{x}_{k-2}$}] (s0) [right = of p1.center] {};
		\node[interpolated] (s1) [right =of s0, xshift=\dx] {};
		\node[interpolated] (s2) [right =of s1, xshift=\dx] {};
		\node[interpolated] (s3) [right =of s2, xshift=\dx] {};
		\node[state] (s4) [right =of s3, xshift=\dx] {};
		\node[interpolated] (s5) [right =of s4, xshift=\dx] {};
		\node[interpolated] (s6) [right =of s5, xshift=\dx] {};
		\node[interpolated] (s7) [right =of s6, xshift=\dx] {};
		\node[state, label=above right:{$\mathbf{x}_{k}$}] (s8) [right =of s7, xshift=\dx] {};
		
		\draw[graphedge] (s0.east) edge node[midway, pfactor] {} (s1.west);
		\draw[graphedge] (s1.east) edge node[midway, pfactor] {} (s2.west);
		\draw[graphedge] (s2.east) edge node[midway, pfactor] {} (s3.west);
		\draw[graphedge] (s3.east) edge node[midway, pfactor] {} (s4.west);
		\draw[graphedge] (s4.east) edge node[midway, pfactor] {} (s5.west);
		\draw[graphedge] (s5.east) edge node[midway, pfactor] {} (s6.west);
		\draw[graphedge] (s6.east) edge node[midway, pfactor] {} (s7.west);
		\draw[graphedge] (s7.east) edge node[midway, pfactor] {} (s8.west);
		
		\node[pfactor] (v0) at ($(s0.center) + (0, \vy)$) {};
		
		\node[vfactor] (v1) at ($(s1.center) + (0, \vy)$) {};
		\node[vfactor] (v2) at ($(s2.center) + (0, \vy)$) {};
		\node[vfactor] (v3) at ($(s3.center) + (0, \vy)$) {};
		\node[vfactor] (v4) at ($(s4.center) + (0, \vy)$) {};
		\node[label={$\mathbf{x}_{k-1}$}]  (ttt) at ($(s4.center) + (0, \vy)$) {};
		\node[vfactor] (v5) at ($(s5.center) + (0, \vy)$) {};
		\node[vfactor] (v6) at ($(s6.center) + (0, \vy)$) {};
		\node[vfactor] (v7) at ($(s7.center) + (0, \vy)$) {};
		\node[vfactor] (v8) at ($(s8.center) + (0, \vy)$) {};
		
		\node[preintfactor] (pi2) at ($(v2.center) + (0, \vy)$) {};
		\node[preintfactor] (pi6) at ($(v6.center) + (0, \vy)$) {};
		\draw[graphedge] (s0.north) edge[bend left=20] (pi2);
		\draw[graphedge] (pi2) edge[bend left=20] (s4.north);
		\draw[graphedge] (s4.north) edge[bend left=20] (pi6);
		\draw[graphedge] (pi6) edge[bend left=20] (s8.north);
		
		\draw[graphedge] (s0) -- (v0);
		\draw[graphedge] (s1) -- (v1);
		\draw[graphedge] (s2) -- (v2);
		\draw[graphedge] (s3) -- (v3);
		\draw[graphedge] (s4) -- (v4);
		\draw[graphedge] (s5) -- (v5);
		\draw[graphedge] (s6) -- (v6);
		\draw[graphedge] (s7) -- (v7);
		\draw[graphedge] (s8) -- (v8);
		
		\node[mappoint] (mm1) at ($(s0.center) + (1mm, -11mm)$) {};
		\node[mappoint] (mm2) at ($(s0.center) + (4mm, -13mm)$) {};
		\node[mappoint] (mm3) at ($(s0.center) + (4mm, -15mm)$) {};
		\node[mappoint] (mm4) at ($(s0.center) + (6mm, -17mm)$) {};
		\node[mappoint] (mm5) at ($(s0.center) + (8mm, -19mm)$) {};
		\node[mappoint] (mm6) at ($(s0.center) + (8mm, -21mm)$) {};
		\node[mappoint] (mm7) at ($(s0.center) + (6mm, -24mm)$) {};
		\node[mappoint] (mm8) at ($(s0.center) + (8mm, -27mm)$) {};
		\node[mappoint] (mm11) at ($(s0.center) + (-2mm, -11mm)$) {};
		\node[mappoint] (mm12) at ($(s0.center) + (0mm, -13mm)$) {};
		\node[mappoint] (mm13) at ($(s0.center) + (2mm, -16mm)$) {};
		\node[mappoint] (mm14) at ($(s0.center) + (5mm, -19mm)$) {};
		\node[mappoint] (mm15) at ($(s0.center) + (2mm, -20mm)$) {};
		\node[mappoint] (mm16) at ($(s0.center) + (-1mm, -18mm)$) {};
		
		\draw[graphedge] (s1.south) edge[bend left=20] node[midway, icpfactor] {} (mm1);
		\draw[graphedge] (s2.south) edge[bend left=20] node[midway, icpfactor] {} (mm2);
		\draw[graphedge] (s3.south west) edge[bend left=20] node[midway, icpfactor] {} (mm3);
		\draw[graphedge] (s4.south west) edge[bend left=20] node[midway, icpfactor] {} (mm4);
		\draw[graphedge] (s5.south west) edge[bend left=20] node[midway, icpfactor] {} (mm5);
		\draw[graphedge] (s6.south west) edge[bend left=20] node[midway, icpfactor] {} (mm6);
		\draw[graphedge] (s7.south west) edge[bend left=20] node[midway, icpfactor] {} (mm7);
		\draw[graphedge] (s8.south west) edge[bend left=20] node[midway, icpfactor] {} (mm8);
		
		
		\coordinate (legend) at ($(s0.west)!0.3!(s8.east) + (0mm, -32mm)$);
		\node[state, left=35mm of legend, label={[font=\footnotesize, label distance=0.5mm]right:Estimated State}] (c2) {};
		\node[dot, black, minimum size=1mm, left=12mm of legend, yshift=0mm, label={[font=\footnotesize, label distance=1mm]right:Local Map Point}] {};
		\node[interpolated, left=-14.5mm of legend, yshift=0mm, label={[font=\footnotesize, label distance=0.5mm]right:Interpolated State}] {};

		\node[dot, red, left=12mm of legend, yshift=-5mm, label={[font=\footnotesize, label distance=0.5mm]right:Motion Prior $J_v$}] {};
		\node[dot, blue, left=38mm of legend, yshift=-5mm, label={[font=\footnotesize, label distance=0.5mm]right:ICP Factor $J_{\text{p2p}}$}] {};
		\node[dot, green, left=-14mm of legend, yshift=-5mm, label={[font=\footnotesize, label distance=0.5mm]right:Gyro Factor $J_{\omega}$}] {};
		\node[dot, gray, left=-38mm of legend, yshift=-5mm, label={[font=\footnotesize, label distance=0.5mm]right:Preintegrated Velocity Factor $J_{a}$}] {};
	\end{tikzpicture}
	\caption{This figure depicts a factor graph of our sliding window lidar-inertial odometry using a continuous-time motion prior. The larger triangles represent the estimation times that form our sliding window. The state $\mathbf{x}(t) = \{\mathbf{T}(t), \boldsymbol{\varpi}(t), \mathbf{b}(t) \}$ includes the pose $\mathbf{T}(t)$, the body-centric velocity $\boldsymbol{\varpi}(t)$, and the IMU biases $\mathbf{b}(t)$. The grey-shaded state $\mathbf{x}_{k-2}$ is next to be marginalized and is held fixed during the optimization of the current window. The smaller triangles are interpolated states that we do not directly estimate during the optimization process. Instead, we construct continuous-time measurement factors using the posterior Gaussian process interpolation formula in Equation~\ref{eq:interp}. The ICP measurement times and gyroscope measurement times may be asynchronous. The preintegrated velocity factors do not need to align with the estimated state times and could be between two interpolated states instead. We include a unary prior on $\mathbf{x}_{k-2}$ to denote the prior information from the sliding window filter.}
	\label{fig:cticp_factor_graph}
\end{figure*}

\begin{equation}
	\boldsymbol{\gamma}(t) = \begin{bmatrix} \ln(\mathbf{T}(t) \mathbf{T}(t_k)^{-1})^\vee \\ \J \left( \ln(\mathbf{T}(t) \mathbf{T}(t_k)^{-1})^\vee \right)^{-1} \boldsymbol{\varpi}(t) \end{bmatrix}.
\end{equation} The motion prior factors can then be written in terms of the global variables,
\begin{equation}
	\mathbf{e}_{v,k} = \begin{bmatrix} \ln(\mathbf{T}_{k+1} \mathbf{T}_k^{-1} )^\vee - (t_{k+1} - t_k) \w_k \\ \J \left( \ln(\mathbf{T}_{k+1} \mathbf{T}_k^{-1} )^\vee \right)^{-1} \w_{k+1} - \w_k \end{bmatrix},
\end{equation} where we observe that the motion prior is penalizing the state estimates from deviating from a constant velocity.

After performing batch trajectory estimation using these motion prior factors, the sparsity of the prior allows Gaussian process interpolation to be performed as an $O(1)$ operation where
\begin{align}
	\mathbf{\hat{T}}(\tau) =~&\exp\left( ( \boldsymbol{\Lambda}_1(\tau) \hat{\boldsymbol{\gamma}}_k(t_k) + \boldsymbol{\Psi}_1(\tau) \hat{\boldsymbol{\gamma}}_k(t_{k+1}) )^\wedge \right) \mathbf{\hat{T}}_k, \nonumber  \\
	\hat{\boldsymbol{\varpi}}(\tau) =~&\J( \ln( \mathbf{\hat{T}}(\tau) \mathbf{\hat{T}}_k^{-1} )^\vee ) \nonumber \\
	&\times ( \boldsymbol{\Lambda}_2(\tau) \hat{\boldsymbol{\gamma}}_k(t_k) + \boldsymbol{\Psi}_2(\tau) \hat{\boldsymbol{\gamma}}_k(t_{k+1}) ) \label{eq:interp}
\end{align} are the interpolation equations involving only the two states bracketing the desired interpolation time: $t_k < \tau < t_{k+1}$. The interpolation matrices that result from the standard GP interpolation formula are
\begin{subequations}
\begin{align}
	\boldsymbol{\Psi}(\tau) &= \mathbf{Q}_\tau \boldsymbol{\Phi}(t_{k+1}, \tau)^T \mathbf{Q}_{k+1}^{-1}, \\
	\boldsymbol{\Lambda}(\tau) &= \boldsymbol{\Phi}(\tau, t_k) - \boldsymbol{\Psi}(\tau) \boldsymbol{\Phi}(t_{k+1}, t_k).
\end{align}
\end{subequations} where $\boldsymbol{\Psi}(\tau) = [\boldsymbol{\Psi}_1(\tau)^T ~ \boldsymbol{\Psi}_2(\tau)^T]^T$ and $\boldsymbol{\Lambda}(\tau) = [\boldsymbol{\Lambda}_1(\tau)^T ~ \boldsymbol{\Lambda}_2(\tau)^T]^T$. When performing continuous-time trajectory estimation, we use the posterior interpolation formulas to build measurement factors at times between our desired estimation times.

\section{Radar-Inertial and Lidar-Inertial Odometry}






In this section, we describe our lidar-inertial odometry, which is implemented as sliding-window batch trajectory estimation. The factor graph corresponding to our approach is depicted in Figure~\ref{fig:cticp_factor_graph}. The state $\mathbf{x}(t) = \{\mathbf{T}(t), \boldsymbol{\varpi}(t), \mathbf{b}(t)\}$ consists of the $SE(3)$ pose $\mathbf{T}_{vi}(t)$, the body-centric velocity $\boldsymbol{\varpi}_v^{vi}(t)$, and the IMU biases $\mathbf{b}(t)$. In our notation, $\boldsymbol{\varpi}_v^{vi}$ is a $6\times1$ vector containing the body-centric linear velocity $\boldsymbol{\nu}_v^{vi}$ and angular velocity $\boldsymbol{\omega}_v^{vi}$. We use a white-noise-on-acceleration prior, as defined in \eqref{eq:wnoa}. Our IMU measurement model is
\begin{equation}
\begin{bmatrix} \tilde{\mathbf{a}} \\ \tilde{\boldsymbol{\omega}} \end{bmatrix} = \begin{bmatrix}  \mathbf{a}_v^{vi} - \mathbf{C}_{vi}\mathbf{g}_i \\ \boldsymbol{\omega}_v^{vi} \end{bmatrix} + \begin{bmatrix} \mathbf{b}_a \\ \mathbf{b}_{\omega} \end{bmatrix} + \begin{bmatrix} \mathbf{w}_a \\ \mathbf{w}_{\omega} \end{bmatrix}
\end{equation} where $\mathbf{b}_a$ and $\mathbf{b}_\omega$ are the accelerometer and gyroscope biases, $\mathbf{w}_a \sim \mathcal{N}(\mathbf{0}, \mathbf{R}_{a})$ and $\mathbf{w}_\omega \sim \mathcal{N}(\mathbf{0}, \mathbf{R}_{\omega})$ are zero-mean Gaussian noise. Due to angular velocity being a part of the state, the associated gyroscope error function is straightforward:
\begin{subequations}
\begin{align} 
	J_{\omega, \ell} &= \frac{1}{2} \mathbf{e}_{\omega,\ell}^T \mathbf{R}_\omega^{-1} \mathbf{e}_{\omega,\ell}, \\
	\mathbf{e}_{\omega,\ell} &= \tilde{\boldsymbol{\omega}}_\ell - \boldsymbol{\omega}(\tau_{\ell}) - \mathbf{b}_{\omega}(\tau_{\ell}). \label{eq:gyro_error}
\end{align}
\end{subequations} We preintegrate accelerometer measurements over a short temporal window $t_k \leq \tau_1 < \cdots < \tau_N < t_{k+1}$ to form a relative velocity factor,
\begin{align}
\Delta \boldsymbol{\nu}(t_{k+1}, \tau_1) = \sum_{n=1}^N \big( \tilde{\mathbf{a}}_n + \mathbf{C}_{vi}(\tau_n) \mathbf{g}_i - \mathbf{b}_a(\tau_n) \big) \Delta t_n, \label{eq:preint}
\end{align} where the associated factor is given by
\begin{subequations}
\begin{align}
	J_{a,k} &= \frac{1}{2}\mathbf{e}_{a,k}^T \mathbf{R}_a(t_{k+1}, \tau_1)^{-1} \mathbf{e}_{a,k}, \\
	\mathbf{e}_{a,k} &= \boldsymbol{\nu}(t_{k+1}) - \boldsymbol{\nu}(\tau_1) - \Delta \boldsymbol{\nu}(t_{k+1}, \tau_1). \label{eq:acc_error}
\end{align}	
\end{subequations} The covariance associated with the preintegrated velocity factor is $\mathbf{R}_a(t_{k+1}, \tau_1)=\sum_n \mathbf{R}_a \Delta t_n^2$. In error functions \eqref{eq:gyro_error} \eqref{eq:acc_error}, we use a continuous-time interpolation of the state.  In order to interpolate for the measurement times, we use the posterior GP interpolation formula \eqref{eq:interp}. Interpolating the state at a given measurement time effectively converts a unary measurement factor into a binary factor between the two bracketing estimation times. For example, a first-order linearization of the gyroscope error is given by
\begin{equation}
	\mathbf{e}_{\omega,\ell} \approx \mathbf{\bar{e}}_{\omega,\ell} + \frac{\partial \mathbf{e}_{\omega,\ell}}{\partial \delta \boldsymbol{\omega}(\tau_\ell)} \left(\frac{\partial \delta \boldsymbol{\omega}(\tau_\ell)}{\partial \delta \mathbf{x}_k} \delta \mathbf{x}_k + \frac{\partial \delta \boldsymbol{\omega}(\tau_\ell)}{\partial \delta \mathbf{x}_{k+1}} \delta \mathbf{x}_{k+1} \right),
\end{equation} where we have included the Jacobians of the perturbation at the interpolated time $\tau_\ell$ with respect to the perturbations at the bracketing estimation times $(t_k, t_{k+1})$. We provide these interpolation Jacobians in the appendix. Using the posterior interpolation formula in this way is an approximation as this is not equivalent to marginalizing out the measurement times. However, we have found this approximation to be fast and to work well in practice. The computational cost of our approach scales linearly with both the number estimation times and the number of measurement times. This is different from the approach of Le Gentil and Vidal-Calleja \cite{legentil_ijrr23} where the computational cost of preintegration scales with the cube of the number of estimation times in the preintegration window.









%

We use point-to-plane factors similar to iterative closest point (ICP). The associated error function is
\begin{subequations}
\begin{align}
	J_{\text{p2p},j} &= \mathbf{e}_{\text{p2p},j}^T \mathbf{R}_{\text{p2p},j}^{-1} \mathbf{e}_{\text{p2p},j}, \\
	\mathbf{e}_{\text{p2p},j} &= \alpha_j \mathbf{n}_j^T \mathbf{D} (\mathbf{p}_j - \mathbf{T}_{vi}(\tau_j)^{-1} \mathbf{T}_{vs} \mathbf{q}_j),
\end{align}
\end{subequations} where $\mathbf{q}_j$ is the query point, $\mathbf{p}_j$ is the matched point in the local map, $\mathbf{n}_j$ is an estimate of the surface normal at $\mathbf{p}_j$ given neighboring points in the map, $\mathbf{D}$ is a constant matrix removing the homogeneous component, $\mathbf{T}_{vs}$ is an extrinsic calibration between the lidar frame and the vehicle frame, and $\alpha = (\sigma_2 - \sigma_3)/\sigma_1$ \cite{dellenbach_icra22, deschaud_icra18} is a heuristic weight to favour planar neighborhoods. Query points are matched to a sliding local voxel map centered on the current estimate of the robot's position. Once a voxel has reached its maximum number of allocated points, new points are not added to it. This helps to keep the state estimate from \change{exhibiting a random walk} while stationary \change{by keeping the map fixed}. Depending on the dataset, we clear voxels in the map if they have not been observed for \change{approximately one second}. We found this to be important in the Boreas dataset, especially for sequences with snowstorms where erroneous snow detections would accumulate in the map and eventually cause \change{ICP to fail by drastically increasing the number of outlier points}. Interestingly, the addition of IMU measurements made our lidar-inertial pipeline more robust to this accumulation of noise.

Our Gaussian process prior introduces a set of motion prior factors between estimation times, which penalize the state for deviating from a constant velocity. These motion prior factors are defined in \eqref{eq:prior1} \eqref{eq:prior2}. We also include motion prior factors for the IMU biases,
\begin{subequations}
\begin{align}
	J_{v,b,k} = \frac{1}{2} \mathbf{e}_{v,b,k}^T \mathbf{Q}_{b,k}^{-1} \mathbf{e}_{v,b,k}, \\
	\mathbf{e}_{v,b,k} = \mathbf{b}(t_{k+1}) - \mathbf{b}(t_{k}),
\end{align}
\end{subequations} where $\mathbf{Q}_{b,k} = \bm{Q}_b \Delta t_k$ is the covariance resulting from a white-noise-on-velocity motion prior, and $\bm{Q}_b$ is the associated power spectral density matrix. The objective function that we seek to minimize is then
\begin{equation}
J = \sum_k J_{v,k} + \sum_j J_{\text{p2p},j} + \sum_\ell J_{\omega,\ell} + \sum_k J_{a,k}.
\end{equation}

We solve this nonlinear least-squares problem for the optimal perturbation to the state using Gauss-Newton. Once the solver has converged, we update the pointcloud correspondences and iterate this two-step process to convergence. In practice, we typically limit the maximum number of inner-loop Gauss-Newton iterations to 5, and the number of outer-loop iterations to 10 in order to enable real-time operation.

\begin{figure*}[t]
	\centering
	\begin{tikzpicture}[
		node distance=0.1cm,
		module/.style={inner sep=0, draw, fill=red!10!white, rectangle, minimum width=2cm, minimum height=8mm, text width=1.5cm, font=\small, align=center,line width=1pt},
		input/.style={inner sep=0, draw, fill=cyan!30!white, rectangle, rounded corners=0.2cm, minimum width=1.5cm, minimum height=0.8cm, text width=1.5cm, font=\small, align=center, line width=1pt},
		output/.style={inner sep=0, draw, fill=blue!10!yellow!20!white, rectangle, rounded corners=0.2cm, minimum width=1.5cm, minimum height=0.8cm, text width=1.5cm, font=\small, align=center, line width=1pt},
		custom arrow/.style={draw, single arrow, minimum height=0.9cm, minimum width=0.9cm, single arrow head extend=0.15cm, line width=1pt, scale=0.25, fill=black},
		rotate border/.style={shape border uses incircle, shape border rotate=#1},
		bounding box wide/.style={yshift=0.25cm, inner sep=0, draw, rectangle, dashed, minimum width=4.6cm, minimum height=4.1cm, text depth = 3.7cm, line width=1pt},
		bounding box thin/.style={bounding box wide, minimum width=2.1cm},
		]
		\node[input, minimum width=2.0cm] (sensor input) at (0,0) {Radar/Lidar};
		\node[custom arrow] (a0) [right=of sensor input] {};
		
		\node[module] (feature extraction) [right=of a0] {Keypoint Extraction};
		\node[custom arrow] (a1) [right=of feature extraction] {};
		\node[module] (filtering) [right=of a1] {Downsample};
		
		\node[custom arrow] (a2) [right=of filtering] {};
		
		\node[module] (registration) [right=of a2] {Optimization};
		
		\node[custom arrow, rotate border=-90] (a23) [above=of registration, xshift=25mm] {};
		\node[custom arrow, rotate border=-90] (a24) [above=of registration, xshift=-25mm] {};
		
		\node[cylinder, draw, shape border rotate=90, minimum width=10mm, aspect=0.5,thick, fill=lightgray] (map) [above=of a23] {map};
		
		\definecolor{lightgreen}{HTML}{6afc6a};
		
		\node[input, minimum width=6mm, fill=green!20!white, inner sep=0pt, text width=10mm] (imuinput) [above=of a24] {IMU};
		
		
		\node[custom arrow, rotate border=-90] (a22) [below=of registration] {};
		\node[output] (trksubmap) [below=of a22] {$\hat{\mathbf{x}}(t)$};
		
		\node[custom arrow] (a3) [right=of registration] {};
		\node[module] (mapmaintenance) [right=of a3] {Map\\Maintenance};
		
		\node[rectangle, minimum width=1.6mm, minimum height=9mm, fill=black, inner sep=0pt, line width=0pt] (rectarrow1) [above=of mapmaintenance] {};
		\node[rectangle, minimum width=10mm, minimum height=1.6mm, fill=black, inner sep=0pt, line width=0pt] (rectarrow2) [above=of rectarrow1, xshift=-5mm, yshift=-2.6mm] {};
		\node[custom arrow, rotate border=180] (a4) [left=of rectarrow2.west, xshift=4mm] {};
		
		
		
		
		%
		%
		%
		
	\end{tikzpicture}
	
	\caption{This figure depicts the simple architecture diagram for STEAM-LIO. In the radar-based pipelines, keypoints are first extracted using a constant false alarm rate (CFAR) detector. In the lidar-based pipelines, we randomly shuffle the order of the points and then downsample using a coarse voxel grid. At the optimization stage, we alternate between finding correspondences between the live undistorted pointcloud and the sliding local map, and estimating the trajectory using sliding window batch trajectory estimation. The inner loop of the optimization stage involves optimizing a nonlinear least-squares problem with Gauss-Newton. At the map maintenance stage, we add registered points to the sliding local map, and optionally cull voxels that have been unobserved for several consecutive frames.}
	\label{fig:architecture_diagram}
\end{figure*}
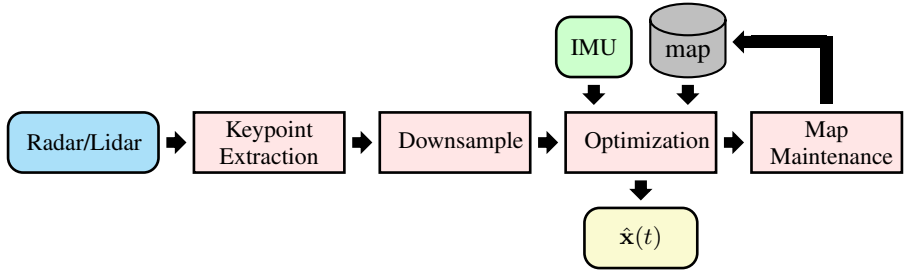

\begin{algorithm}[t]
	\caption{STEAM-LIO}\label{alg:steamlio}
	\begin{algorithmic}[1]
		\INPUT map: $\{\mathbf{p}_i\}$, new frame: $\{\mathbf{q}_j, \tau_j\}$, IMU: $\{\tilde{\boldsymbol{\omega}}_\ell, \tilde{\mathbf{a}}_\ell\},$
		\Statex $\quad\quad\hat{\mathbf{x}}(t), \mathbf{A}, \mathbf{c}$ from previous iteration 
		\OUTPUT $\hat{\mathbf{x}}(t) = \{\hat{\mathbf{T}}(t), \hat{\w}(t), \hat{\mathbf{b}}(t)\}$ where $t \in [t_{k-2}, t_k]$
		\newline
		\State $\mathbf{T}(t_k) \gets \exp(\Delta t_k \w(t_{k-1})^\wedge) \mathbf{T}(t_{k-1})$ \\ $\w(t_k) \gets \w(t_{k-1})$, $\mathbf{b}(t_k) \gets \mathbf{b}(t_{k-1})$
		\State $\hat{\mathbf{x}}(t), \mathbf{A}, \mathbf{c} \gets \text{SlideWindow}(\hat{\mathbf{x}}(t), \mathbf{A}, \mathbf{c}, \mathbf{x}_k)$
		\State $\{\mathbf{q}_j, \tau_j\} \gets \text{Downsample}(\{\mathbf{q}_j, \tau_j\})$
		\State $x$ $\gets 0$, $||\Delta \mathbf{x}|| \gets \infty$
		\While{$||\Delta \mathbf{x}|| > T_{\text{outer}} \wedge x < N_{\text{outer}}$}
		\State $\{\bar{\mathbf{q}}_j\} \gets \text{Undistort}(\{\mathbf{q}_j, \tau_j\}, \hat{\mathbf{x}}(t))$
		\State $\{\mathbf{p}_j, \mathbf{n}_j\} \gets \text{Matching}(\{\mathbf{p}_i\}, \{\bar{\mathbf{q}}_j\})$
		\State $J \gets J_{v}(\mathbf{x}_{k-1}, \mathbf{x}_k) + J_\omega(\mathbf{x}(t), \{\tilde{\boldsymbol{\omega}}_\ell\})$
		\Statex $\quad~ + J_a(\mathbf{x}(t), \{\tilde{\mathbf{a}}_\ell\}) + J_\text{p2p}(\mathbf{x}(t), \{\mathbf{p}_j, \mathbf{n}_j, \mathbf{q}_j, \tau_j\}) $
		\State $y \gets 0, ||\delta \mathbf{x}|| \gets \infty, \Delta J \gets \infty, \mathbf{x}_\text{prev} \gets \hat{\mathbf{x}}(t)$
		\While {$||\delta \mathbf{x}|| > T_{\text{inner}} \wedge \Delta J > \delta J \wedge y < N_\text{inner}$}
		\State $\mathbf{A}, \mathbf{c} \gets \text{buildAndUpdateGN}(\mathbf{A}, \mathbf{c}, J, \hat{\mathbf{x}}(t))$
		\State $\delta \mathbf{x} \gets \text{CholeskySolve}(\mathbf{A}, \mathbf{c})$
		\State $\hat{\mathbf{x}}(t) \gets \text{UpdateState}(\hat{\mathbf{x}}(t), \delta \mathbf{x})$
		\State $J, \Delta J \gets \text{UpdateCost}(\hat{\mathbf{x}}(t), J)$
		\State $y \gets y + 1$
		\EndWhile
		\State $||\Delta \mathbf{x}|| \gets \text{Dist}(\mathbf{x}_\text{prev},  \hat{\mathbf{x}}(t))$
		\State $x \gets x + 1$
		\EndWhile
		\State $\{\mathbf{p}_i\} \gets \text{UpdateMap}(\{\mathbf{p}_i\}, \{\mathbf{q}_j, \tau_j\}, \hat{\mathbf{x}}(t))$
	\end{algorithmic}
\end{algorithm}

\subsection{Implementation Details}

Algorithm~\ref{alg:steamlio} provides pseudocode for STEAM-LIO at a high level and Figure~\ref{fig:architecture_diagram} depicts the software architecture for our approach. For a new lidar frame, we first initialize the new state using constant velocity. We then slide the estimation window forward and marginalize out states that are no longer involved in the current optimization problem. For lidar odometry, a coarse voxelization of the input pointcloud is then performed where the default is 1.5m. At each iteration of the outer loop, we first undistort the lidar frame using the posterior trajectory estimate of the previous iteration in order to obtain correspondences between the live frame and the local map. We then build the optimization problem given the set of lidar factors, IMU measurements, and motion prior factors derived from the Gaussian process motion. This optimization problem is then minimized using Gauss-Newton. Finally, the undistorted points are added to the sliding local map. The local map also has a coarse discretization of 1.0m but we allow up to 20 points in each voxel with a minimum point distance of 0.1m. This voxelization strategy is inspired by CT-ICP \cite{dellenbach_icra22}.

In order to achieve real-time performance, we found it necessary to implement timestamp binning where the original timestamp frequency is downsampled to reduce the number of state interpolations and associated Jacobians that need to be computed. For the KITTI-raw and Newer College Dataset, we downsample lidar timestamps to 5kHz. For the Boreas dataset, we downsample lidar timestamps to 400Hz. We have found that by reducing the timestamp frequency in this way, we can retain most of the benefits of continuous-time state estimation while still operating efficiently.

\subsection{Gravity Vector Orientation}

In our approach, we estimate the orientation of the gravity vector relative to the initial map frame at startup. We do this by first estimating the orientation of the gravity frame using an initial set of accelerometer measurements,
\begin{subequations}
\begin{align}
	J &= \sum_n \mathbf{e}_{a,n}^T \mathbf{R}_a^{-1} \mathbf{e}_{a,n} + \ln(\mathbf{C}_{ig})^{\vee^T} \mathbf{\check{P}}^{-1}_g \ln(\mathbf{C}_{ig})^{\vee}\\
	&~~~+ \mathbf{b}_a^T \mathbf{\check{P}}^{-1}_b \mathbf{b}_a, \nonumber \\
	\mathbf{e}_{a,n} &= \mathbf{\tilde{a}}_n - \mathbf{C}_{ig} \mathbf{g} - \mathbf{b}_a,
\end{align}
\end{subequations} where we assume that the robot is stationary at startup and we impose a weak prior on $\mathbf{C}_{ig}$ to constrain the rotational degree of freedom not observed by the accelerometer measurements. This estimate of the gravity vector orientation then serves as a prior for the gravity vector orientation included in the state at $t = 0$. We hold our estimate of the gravity vector orientation fixed once it has been marginalized from the sliding window.

We experimented with including the gravity vector orientation in the state: $\mathbf{x}(t) = \{\mathbf{T}(t), \boldsymbol{\varpi}(t), \mathbf{b}(t), \mathbf{C}_{ig}(t)\}$. In this case, we include a motion prior factor for the gravity vector orientation,
\begin{subequations}
\begin{align}
	J_{v,g,k} &= \frac{1}{2}\mathbf{e}_{v,g,k}^T \mathbf{Q}_{g,k}^{-1}  \mathbf{e}_{v,g,k}, \\
	\mathbf{e}_{v,g,k} &= \ln(\mathbf{C}_{ig}(t_k)\mathbf{C}_{ig}(t_{k+1})^{-1})^{\vee},
\end{align}
\end{subequations}


In our experiments, we did not observe any benefit from including the gravity vector orientation in the state. However, some recent work by Nemiroff et al. \cite{nemiroff_iros23} has shown that it can be beneficial to mapping accuracy in challenging scenarios.

\subsection{Sliding Window Marginalization}

In our approach, we perform sliding window batch trajectory estimation. The length of the sliding window is equivalent to two lidar frames or roughly 200ms. We output the pose at the middle of the newest lidar frame so that the latency is equivalent to competing approaches. In Figure~\ref{fig:cticp_factor_graph}, the darkly shaded state $\mathbf{x}_{k-2}$ is slated to be marginalized and is held fixed during optimization. However, there are still several continuous-time measurement factors between states $\mathbf{x}_{k-2}$ and $\mathbf{x}_{k-1}$. As such, at each iteration of our Gauss-Newton solver, we first interpolate the state at each measurement time and update the associated measurement Jacobians before marginalizing $\mathbf{x}_{k-2}$. For example,
\begin{align}
\begin{bmatrix}
	\mathbf{A}_{k-2,k-2} & \mathbf{A}_{k-1, k-2}^T & \\
	\mathbf{A}_{k-1, k-2} & \mathbf{A}_{k-1, k-1} & \mathbf{A}_{k, k-1}^T \\
	& \mathbf{A}_{k, k-1} & \mathbf{A}_{k, k}
\end{bmatrix}
\begin{bmatrix}
	\delta \mathbf{x}_{k-2}^\star \\ \delta \mathbf{x}_{k-1}^\star \\ \delta \mathbf{x}_{k}^\star
\end{bmatrix}
= 
\begin{bmatrix}
	\mathbf{c}_{k-2} \\ \mathbf{c}_{k-1} \\ \mathbf{c}_{k}
\end{bmatrix}
\end{align} becomes
\begin{align}
	& \left[ \begin{matrix}
		\mathbf{A}_{k-1, k-1} - \mathbf{A}_{k-1, k-2} \mathbf{A}_{k-2,k-2}^{-1} \mathbf{A}_{k-1, k-2}^T\\
		\mathbf{A}_{k, k-1} 
	\end{matrix} \right. \cdots \\
	&\left. \begin{matrix} \mathbf{A}_{k, k-1}^T \\ \mathbf{A}_{k, k} \end{matrix} \right]
	\begin{bmatrix}
		\delta \mathbf{x}_{k-1}^\star \\ \delta \mathbf{x}_{k}^\star
	\end{bmatrix} 
	= \begin{bmatrix}
		\mathbf{c}_{k-1} - \mathbf{A}_{k-1, k-2} \mathbf{A}_{k-2,k-2}^{-1} \mathbf{c}_{k-2} \\ \mathbf{c}_{k}
	\end{bmatrix}. \nonumber
\end{align}

\begin{figure*}[ht]
	\centering
	\includegraphics[width=1.0\textwidth]{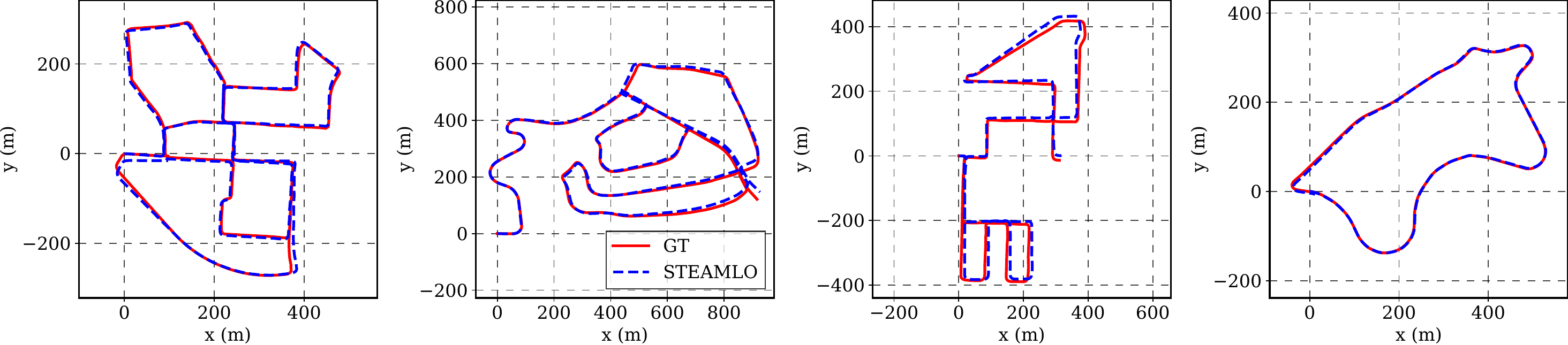}
	\caption{Odometry results on the KITTI-raw dataset. Sequences shown from left to right are 00, 02, 07, 08. The ground-truth trajectory is shown as a solid red line and the STEAM-LO trajectory estimate is shown as a blue dashed line. Note that the estimated trajectory was computed in an online fashion with a sliding window of 200ms and does not make use of any loop closures.}
	\label{fig:kitti_raw_odom}
\end{figure*}



\subsection{Radar-Inertial Odometry}

The architecture of our radar-inertial odometry is largely the same as our lidar-inertial odometry. We note the important differences here. The radar that we use in this work is the Navtech CIR304-H, a mechanical spinning radar that provides a $360^\circ$ horizontal field of view. This sensor is 2D only, as such we do not estimate the orientation of the gravity vector and we remove the gravity term from the preintegrated velocity factor in \eqref{eq:preint}. The Navtech sensor outputs a raw polar radar image corresponding to a power vs. range spectrum for each scanned azimuth. We use a constant false alarm rate (CFAR) detector with an additional constant threshold tuned to the noise floor of the sensor \cite{alhashimi_arxiv21} where we retain the maximum of the left and right subwindows, a variant known as GO-CFAR \cite{rohling1983radar}. The output of CFAR detection is a pointcloud that we register to a sliding local map. When a voxel in the map has not been observed for several consecutive frames (one second), we delete the points in this voxel. Without this map maintenance procedure, we have found that our radar odometry has a tendency to fail due to the significant amount of noise present in the radar pointclouds. The Navtech radar scans each azimuth only once and as such range measurements are corrupted by a Doppler distortion dependent on the robot's egomotion \cite{burnett_ral21}. The Doppler-compensated point-to-point ICP error for radar odometry is then
\begin{subequations}
\begin{align}
	&\mathbf{e}_{\text{p2p},j} = \rho\left( \mathbf{D} \left( \mathbf{p}_j - \mathbf{T}(\tau_j)^{-1} \mathbf{T}_{vs} (\mathbf{q}_j  + \Delta \mathbf{q}_j) \right) \right), \\
	&\text{where}~\Delta \mathbf{q}_j = \beta \mathbf{D}^T  \mathbf{a}_j \mathbf{a}_j^T \mathbf{D} \mathbf{q}_j^\odot \T_{sv} \boldsymbol{\varpi}(\tau_j),
\end{align}
\end{subequations} and where $\rho(\cdot)$ is a Cauchy robust cost function, $\mathbf{D}$ is constant matrix that removes the homogeneous component, $\mathbf{p}_j$ is a reference point in the local map, $\mathbf{T}(\tau_j)$ is the continuous-time interpolation of the robot pose, $\mathbf{T}_{vs}$ is an extrinsic transformation between the sensor and vehicle frame, $\mathbf{q}_j$ is the live query point, $\Delta \mathbf{q}_j$ is an additive correction factor to compensate for the Doppler distortion, $\beta$ is Doppler distortion constant inherent to the sensor \cite{burnett_ral21}, $\mathbf{a}_j$ is a $3 \times 1$ unit vector in the direction of $\mathbf{q}_j$, the $\odot$ operator swaps the order of the operands associated with the skew-symmetric $\wedge$ operator \cite{barfoot_ser17}, and $\T_{sv} = \text{Ad}(\mathbf{T}_{sv})$.


\section{Experimental Results}

We provide experimental results on three datasets, KITTI-raw \cite{geiger_ijrr13}, the Newer College Dataset (NCD) \cite{ramezani_iros20}, and the Boreas dataset \cite{burnett_ijrr23}. KITTI-raw was chosen as it is a popular dataset for benchmarking lidar odometry. The raw version of the dataset contains the motion-distorted pointclouds whereas the original version of the dataset is motion-compensated using GPS poses. Since the purpose of this work is to demonstrate continuous-time state estimation using motion-distorted sensors, we present results for \change{KITTI-raw and not the original KITTI dataset}. The Newer College dataset was chosen as it has become a \change{standard} dataset for benchmarking lidar-inertial odometry. The \change{NCD} dataset is somewhat unique in that \change{it} features several sequences with aggressive high-frequency motions obtained using a handheld sensor mast. These types of trajectories are rarely observed when working with heavy ground robots. Finally, we provide results using the Boreas dataset in order to demonstrate continuous-time radar inertial odometry and to provide a detailed comparison with lidar. In all experiments, we provide average runtime estimates using an Intel Xeon CPU E5-2698 v4 \change{with} 16 threads.

\change{For the lidar-based pipelines, we use the same parameters for} the diagonal of $\bm{Q}$, the power spectral density matrix, $\text{diag}(\bm{Q}) = \{50,50,50,5,5,5\}$. These parameters were obtained by tuning on a training split of the Boreas dataset and were verified to work well on KITTI-raw \change{and the Newer College Dataset}. The diagonal of $\bm{Q}^{-1}$ \change{can be understood as weighting cost terms on body-centric acceleration.} We tune the IMU measurement covariances and bias motion priors for Boreas and the Newer College Dataset separately. \change{We downsample lidar timestamps to 400Hz on the Boreas dataset in order to achieve real-time performance. However, we do not downsample timestamps on KITTI-raw or the Newer College Dataset as we were already able to run in real-time.} On the Boreas dataset, we clear voxels that have not been observed for one second. On the Newer College Dataset, we incrementally build a map to enable implicit loop closures by revisiting previously mapped areas.



\begin{table*}[ht]
	\centering
	\caption{KITTI-raw results (22km / 0.6h): KITTI RTE (\%). The average is computed over all segments of all sequences as in \cite{dellenbach_icra22}. Note that CT-ICP optimizes one lidar frame at a time, while our algorithm optimizes two frames in a sliding window. For a fair comparison, we evaluate our algorithm using the estimated poses at the front of the window (i.e., newest timestamp).}
	\label{tab:kitti-quantitative}
	\begin{tabular}{ l ? c c c c c c c c c c c ? c ? c ? l}
		\toprule
		\midrule
		\textbf{KITTI-raw}              & 00            & 01            & 02            & 03 (N/A)       & 04            & 05            & 06            & 07            & 08            & 09            & 10            & \color{blue}{\textbf{Overall}} & \color{blue}{\textbf{Seq. Avg.}} & $\Delta T$ \\
		\midrule
		CT-ICP \cite{dellenbach_icra22} & 0.51          & 0.81          & 0.55          &              & 0.43          & 0.27          & 0.28 & 0.35          & \textbf{0.80} & 0.47          & \textbf{0.49} & \color{blue}{0.55} &\color{blue}{0.50} & 65ms \cite{dellenbach_icra22}         \\
		\change{KISS-ICP} \cite{vizzo_ral23} & \change{0.51} & \change{0.71} & \change{0.54} && \change{\textbf{0.35}}& \change{0.31}& \change{\textbf{0.26}} & \change{\textbf{0.32}} & \change{0.83} & \change{0.49} & \change{0.58}& \color{blue}{0.55} & \color{blue}{0.49} & \change{26ms} \cite{vizzo_ral23} \\
		STEAM-ICP \cite{wu_ral22} & \textbf{0.49} & 0.65 & \textbf{0.50} &              & 0.38 & \textbf{0.26} & 0.28 & \textbf{0.32} & 0.81          & \textbf{0.46} & 0.53          & \color{blue}{\textbf{0.52}} & \color{blue}{\textbf{0.47}} & 138ms \\
		\change{Constant Velocity} & \change{0.60} & \change{1.62} & \change{0.60} & & \change{0.36} & \change{0.30} & \change{0.27} & \change{0.37} & \change{0.92} & \change{0.52} & \change{0.90} & \color{blue}{0.66} & \color{blue}{0.65} & \change{44ms} \\
		STEAM-LO (Ours) & \textbf{0.49} & \textbf{0.63} & 0.51 &  & 0.38 & \textbf{0.26} & 0.30 & 0.33 & 0.84 & 0.49 & \textbf{0.49} & \color{blue}{0.53} & \color{blue}{\textbf{0.47}} & 89ms\\
		\midrule
		\bottomrule
	\end{tabular}
\end{table*}

\begin{table*}[ht]
	\centering
	\caption{Newer College dataset results (6km / 1.3h): root mean squared ATE (m). Trajectories are aligned with the ground truth using the Umeyama algorithm. $\star$ uses explicit loop closures, $\dagger$ results obtained from \cite{chen_arxiv23}, $\ddagger$ uses camera.}
	\label{tab:newer_college}
	\begin{tabular}{ l ? c c c c c ? l}
		\toprule
		\midrule
		\textbf{Newer College Dataset}              & 01-Short            & 02-Long            & 05-Quad w/ Dynamics            & 06-Dynamic Spinning       & 07-Parkland Mound            & $\Delta T$ \\
		\midrule
		CT-ICP$^\star$ \cite{dellenbach_icra22} & 0.36          &           &          &              &     &  430ms   \cite{dellenbach_icra22}         \\
		\change{KISS-ICP$^\dagger$} \cite{vizzo_ral23} & \change{0.6675} & \change{1.5311} & \change{0.1040} & \change{Failed} & \change{0.2027} & \change{167ms$^\dagger$} \\
		FAST-LIO2$^\dagger$ \cite{xu_tro22} & 0.3775 & 0.3324 & 0.0879 & 0.0771 & 0.1483 & 43ms$^\dagger$ \\
		DLIO \cite{chen_icra23} & 0.3606 & \textbf{0.3268} & \textbf{0.0837} & \textbf{0.0612} & \textbf{0.1196} & 36ms \cite{chen_icra23} \\
		SLICT$^\star$ \cite{nguyen_ral23} & 0.3843 & 0.3496 & 0.1155 & 0.0844 & 0.1290 &  \\
		CLIO$^{\star\ddagger}$ \cite{lv_tmech23} & 0.408 & 0.381 &  & 0.091 &  &  \\
		\change{Constant Velocity} & \change{0.8558} & \change{2.5792} & \change{0.3575} & \change{Failed} & \change{0.5960} & \change{163ms} \\
		STEAM-LO (Ours) & \change{0.3398} & \change{0.4546}  &  \change{0.1083} & \change{0.0802}  & \change{0.1537} & \change{138ms} \\
		STEAM-LO + Gyro (Ours) & \change{0.3055} & \change{0.3340} & \change{0.1090} & \change{0.0824} & \change{0.1444} & \change{76ms} \\
		STEAM-LIO (Ours) & \change{\textbf{0.3042}} & \change{0.3372} & \change{0.1086} & \change{0.0821} & \change{0.1444} & \change{74ms} \\
		\midrule
		\bottomrule
	\end{tabular}
\end{table*}

\begin{table*}[ht]
	\centering
	\footnotesize
	\caption{Boreas Odometry Results (102km / 4.3h): translational drift (\%) / rotational  drift (deg/100m). The first three columns are evaluated in SE(3) whereas the last four columns are evaluated in SE(2).}
	\begin{tabular}{ l ? c c c ? c c c c}
		\toprule
		\midrule
		\textbf{Boreas} &    VTR3-Lidar \cite{burnett_ral22}  & STEAM-LO & STEAM-LIO & STEAM-LO(SE2) & VTR3-Radar \cite{burnett_ral22} & STEAM-RO & STEAM-RIO     \\
		\midrule
		2020-12-04    & 0.49 / 0.14 & 0.41 / 0.13 & 0.39 / 0.13 & 0.13 / 0.05 & 1.92 / 0.53 & \change{1.43 / 0.41} & \change{0.93 / 0.26} \\
		2021-01-26    & 0.51 / 0.16 & 0.62 / 0.21 & 0.53 / 0.18 & 0.30 / 0.11 & 2.27 / 0.66 & \change{1.10 / 0.33} & \change{0.61 / 0.18} \\
		2021-02-09    & 0.49 / 0.14 & 0.38 / 0.13 & 0.38 / 0.13 & 0.14 / 0.06 & 1.94 / 0.59 & \change{1.27 / 0.38} & \change{0.63 / 0.20} \\
		2021-03-09    & 0.57 / 0.17 & 0.47 / 0.15 & 0.46 / 0.15 & 0.13 / 0.05 & 2.00 / 0.59 & \change{1.24 / 0.35} & \change{0.71 / 0.19} \\
		2020-04-22    & 0.49 / 0.15 & 0.39 / 0.13 & 0.39 / 0.13 & 0.13 / 0.05 & 2.56 / 0.63 & \change{1.48 / 0.41} & \change{0.99 / 0.27} \\
		2021-06-29-18    & 0.58 / 0.17 & 0.48 / 0.16 & 0.48 / 0.16 & 0.14 / 0.06 & 1.86 / 0.56 & \change{1.55 / 0.46} & \change{1.04 / 0.29} \\
		2021-06-29-20    & 0.62 / 0.18 & 0.52 / 0.17 & 0.52 / 0.17 & 0.16 / 0.06 & 1.94 / 0.59 & \change{1.70 / 0.48} & \change{0.96 / 0.26} \\
		2021-09-08    & 0.57 / 0.17 & 0.47 / 0.16 & 0.47 / 0.16 & 0.16 / 0.06 & 1.88 / 0.57 & \change{2.01 / 0.59} & \change{1.22 / 0.35} \\
		2021-09-09    & 0.63 / 0.19 & 0.52 / 0.18 & 0.55 / 0.19 & 0.20 / 0.06 & 1.98 / 0.60 & \change{2.16 / 0.64} & \change{1.19 / 0.33} \\
		2021-10-05    & 0.59 / 0.17 & 0.50 / 0.16 & 0.49 / 0.16 & 0.16 / 0.06 & 2.87 / 0.78 & \change{2.27 / 0.63} & \change{1.01 / 0.28} \\
		2021-10-26    & 0.48 / 0.14 & 0.40 / 0.14 & 0.38 / 0.13 & 0.14 / 0.06 & 1.89 / 0.53 & \change{1.88 / 0.53} & \change{0.97 / 0.27} \\
		2021-11-06    & 0.50 / 0.15 & 0.40 / 0.14 & 0.41 / 0.14 & 0.15 / 0.06 & 1.24 / 0.34 & \change{1.86 / 0.54} & \change{1.07 / 0.29} \\
		2021-11-28    & 0.46 / 0.14 & 0.41 / 0.14 & 0.37 / 0.13 & 0.15 / 0.06 & 1.24 / 0.38 & \change{1.95 / 0.57} & \change{1.04 / 0.29} \\
		\midrule
		\color{blue}{\textbf{Seq. Avg.}} & \color{blue}{0.54 / 0.16} & \color{blue}{0.46 / 0.15} & \color{blue}{0.45 / 0.15} & \color{blue}{0.16 / 0.06} & \color{blue}{2.02 / 0.58} & \color{blue}{1.68 / 0.49} & \color{blue}{0.95 / 0.27} \\
		\midrule
		$\Delta T$  & 250ms & 88ms & 97ms & 88ms & 75ms & \change{115ms} & \change{139ms} \\
		\bottomrule
	\end{tabular}
	\label{tab:boreas_quantitative}
\end{table*}


\subsection{KITTI-Raw Results}

The KITTI dataset was collected in Karlsruhe, Germany using an autonomous driving platform equipped with a 64-beam Velodyne lidar and an OXTS RTK GPS. The dataset was primarily collected in an urban environment with some sequences including a brief highway portion. Table~\ref{tab:kitti-quantitative} shows our quantitative results. We compare ourselves against CT-ICP \change{and KISS-ICP}, \change{which represent the state of the art on this dataset}. We also compare ourselves against our previously published work, STEAM-ICP \cite{wu_ral22}. \change{We include two different methods for aggregating the results across sequuences. In the \textit{Overall} column, we concatenate the subsequence errors of the KITTI metric and average across these. In the \textit{Sequence Error} column, we simply average the results for each sequence.} The results show that our translational drift is slightly lower than CT-ICP \change{and KISS-ICP} but not quite as good as STEAM-ICP. However, our approach is demonstrably real-time whereas STEAM-ICP is not. Figure~\ref{fig:kitti_raw_odom} provides some qualitative examples of the trajectories estimated by our lidar odometry. This dataset does not provide raw IMU measurements, as such we cannot use it to benchmark our lidar-inertial odometry. The main purpose of testing on this dataset is to show that, without an IMU, our implementation of continuous-time lidar odometry is competitive with the state of the art.



\subsection{Newer College Dataset Results}

The Newer College Dataset was collected using a handheld sensor mast at the University of Oxford. The dataset includes approximately 6km or 1.3h of data. The sensor suite includes a 64-beam Ouster lidar and an Intel Realsense camera. Both the Ouster lidar and the Intel camera have internal IMUs. We use the 100Hz IMU measurements provided by the Ouster so that we can avoid potential time synchronization problems between the lidar and the IMU. Ground truth for this dataset was obtained by registering lidar scans in the dataset to a surveyed lidar map of the university campus. Table~\ref{tab:newer_college} shows our quantitative results for this dataset. Again, we include CT-ICP \change{and KISS-ICP} since \change{they are} well-known lidar odometry \change{approaches}. DLIO \cite{chen_icra23} and FAST-LIO2 \cite{xu_tro22} are also included as these approaches currently represent the state of the art for lidar-inertial odometry. Finally, we include SLICT \cite{nguyen_ral23} and CLIO \cite{lv_tmech23} as these are continuous-time approaches that use linear interpolation and B-splines, respectively. It is challenging to make a direct comparison to other methods due to \change{significant} difference in front-end preprocessing and map storage strategies. Nevertheless, we show that our approach is competitive with the state of the art while still being real-time capable. We note that, as of writing, ours is the only continuous-time lidar-inertial odometry with confirmed real-time performance on the Newer College Dataset. SLICT quotes their average runtime as 205ms on the NTU Viral dataset \cite{nguyen_ijrr22} using two 16-beam lidars and CLIO quotes their runtime as being 218s for a 397s sequence using a single 16-beam lidar. Note that Table~\ref{tab:newer_college} contains originally published results, except for \change{the results provided for FAST-LIO2 and KISS-ICP} where the results were obtained from \cite{chen_arxiv23}.

\begin{figure}[t]
	\centering
	\includegraphics[width=1.0\columnwidth]{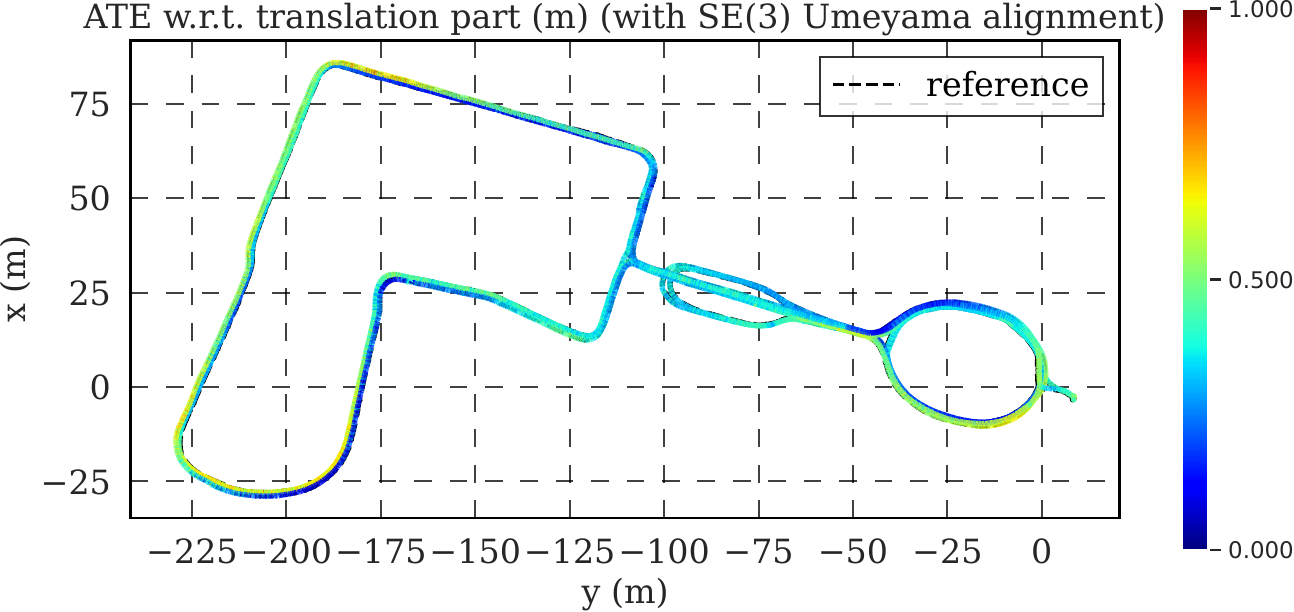}
	\caption{This figure depicts the trajectory estimated by STEAM-LIO during the long sequence of the Newer College Dataset. The trajectory is colored according to the absolute trajectory error when compared against the ground truth. The estimated trajectory was aligned with the ground truth using the Umeyama algorithm \cite{umeyama_tpami91}.}
	\label{fig:ncd_02}
\end{figure}

\begin{figure}[t]
	\centering
	\subfigure[Panoramic image of the courtyard at New College, Oxford]{\includegraphics[width=1.0\columnwidth]{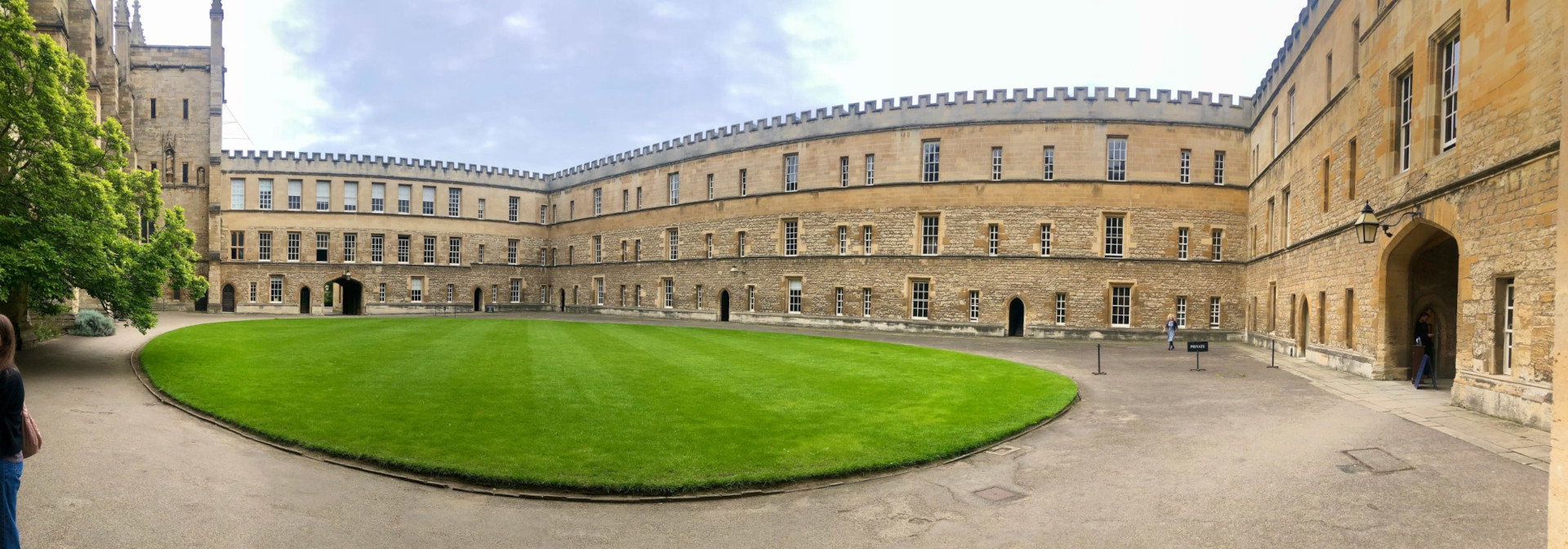}}
	\subfigure[Lidar map of the courtyard colored by reflectivity]{\includegraphics[width=1.0\columnwidth]{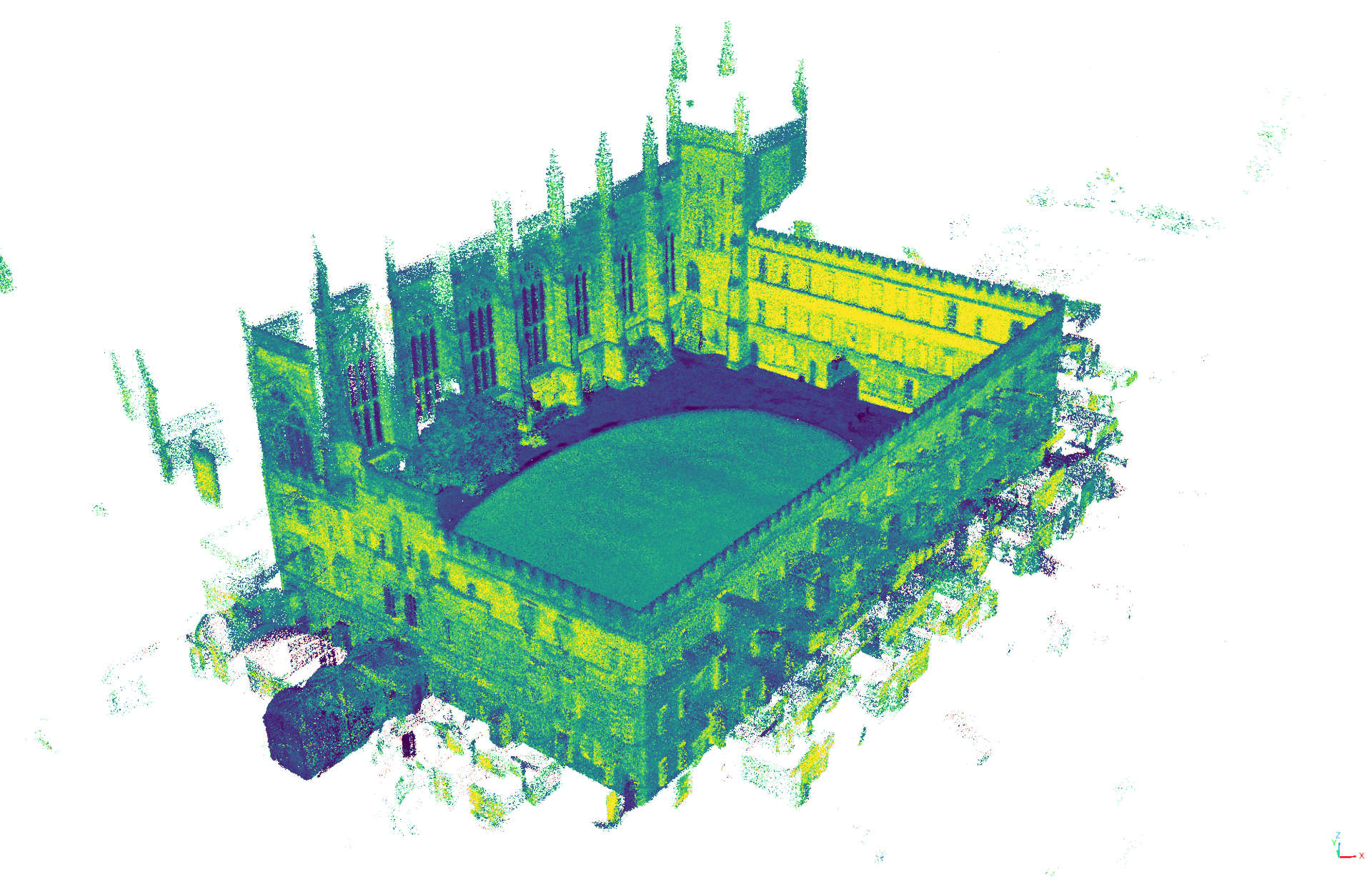}}
	\caption{In this figure, we provide a qualitative example of the map produced by our lidar-inertial odometry using the first 150 seconds of the ``quad with dynamics" sequence from the Newer College Dataset \cite{ramezani_iros20}. In order to produce this figure, we adjusted configuration parameters to produce a denser map.}
	\label{fig:quad_map1}
\end{figure}

\change{Interestingly, the sequences with the most aggressive motions (05, 06) displayed the smallest differences between our continuous-time lidar-only and lidar-inertial odometry. We postulate that this was due to these sequences being collected in a rectangular quad (shown in Figure~\ref{fig:quad_map1}) with plenty of geometric features for point-to-plane ICP. In this dataset, the addition of an IMU seems to have the most noticeable improvement in sequences where there are brief periods lacking sufficient geometric features.} The majority \change{of the lidar-inertial performance improvement} seems to come from using the gyroscope with only a minor additional improvement when the accelerometer is included.

\change{Our lidar-only approach, STEAM-LO achieves better results on sequence 01-Short than all of the previous lidar-inertial methods. Furthermore, STEAM-LO performs the best out of the lidar-only approaches on sequence 06-Dynamic Spinning, where KISS-ICP and our constant-velocity baseline fail due to the rapid rotations observed in this sequence. It can be seen in Table~\ref{tab:newer_college_varying_qc} that our continuous-time approach, STEAM-LO, significantly outperforms KISS-ICP, which relies on a constant-velocity assumption. Our lidar-inertial approach, STEAM-LIO achieves the best performance on sequence 01-Short by a significant margin while remaining competitive with the other lidar-inertial approaches on the other sequences. In fact, if we compute an overall absolute trajectory error for the entire Newer College Dataset by concatenating the squared errors across all timestamps of all sequences, our approach (0.2946m) actually outperforms FAST-LIO2 (0.3152m) and DLIO (0.3048m). The performance of our approach on the Newer College Dataset highlights the value of our continuous-time technique.}


In Figure~\ref{fig:ncd_02}, we provide a qualitative example of the trajectory estimated by our approach. In this case, the trajectory is colored by the absolute trajectory error (ATE). The estimated trajectory is first aligned with the ground truth using the Umeyama algorithm \cite{umeyama_tpami91} before computing the ATE as described in \cite{sturm_iros12}. Even though we do not make use of explicit loop closure factors, we rely on implicitly closing the loop when we revisit previously mapped areas. This allows us to achieve a low ATE for an odometry method. Usually, ATE is used to benchmark SLAM approaches and not odometry. 

We provide another qualitative example of our lidar-inertial odometry in Figure~\ref{fig:quad_map1} where we plot the lidar map generated by our approach alongside a panoramic image of the courtyard of New College, Oxford. In this case, the pointcloud is colored using the Ouster reflectivity. We used a finer voxelization to produce a denser map here. This map was produced using the \textit{Quad with Dynamics} sequence from the Newer College Dataset, which features dynamic swinging motions of the sensor mast. Even with this dynamic motion, we are able to produce a crisp high-quality map.

\change{

\subsection{Ablation Study}



In addition to the continuous-time odometry methods presented in this work, we also present baseline results using a constant-velocity assumption. We approximate the body-centric velocity using the following formula, which includes the poses at the two previous timesteps, \begin{equation}
	\check{\boldsymbol{\varpi}}_{k} \approx \frac{1}{\Delta t_{k-1}} \ln \left( \mathbf{T}_{k-1} \mathbf{T}_{k-2}^{-1} \right)^\vee.
\end{equation}Using this prediction of the velocity, we can deskew the pointcloud using the following formula,
\begin{equation}
	\tilde{\mathbf{q}}_j  := \exp( (t_k - \tau_j) 	\check{\boldsymbol{\varpi}}_{k}^\wedge) \mathbf{T}_{vs} \mathbf{q}_j  .
\end{equation}We then reformulate our point-to-plane error function as
\begin{equation}
		\mathbf{e}_{\text{p2p},j} = \alpha_j \mathbf{n}_j^T \mathbf{D} (\mathbf{p}_j - \mathbf{T}_k^{-1} \tilde{\mathbf{q}}_j ),
\end{equation} where we minimize a cost function including these point-to-plane factors using our nonlinear least-squares solver without including any explicit prior on $\mathbf{T}_k$. We refer to this approach as \textit{Constant Velocity} in Table~\ref{tab:kitti-quantitative} and Table~\ref{tab:newer_college}. On the KITTI-raw dataset, the constant-velocity approach achieves respectable results but is not quite competitive with the state of the art. However, the approach is computationally efficient. On the Newer College Dataset, the gap between the constant-velocity approach and our continuous-time approach is much more apparent. Similar to KISS-ICP, our constant-velocity baseline fails on the Dynamic Spinning sequence.

 \begin{figure}[t]
 	\centering
 	\includegraphics[width=0.8\columnwidth]{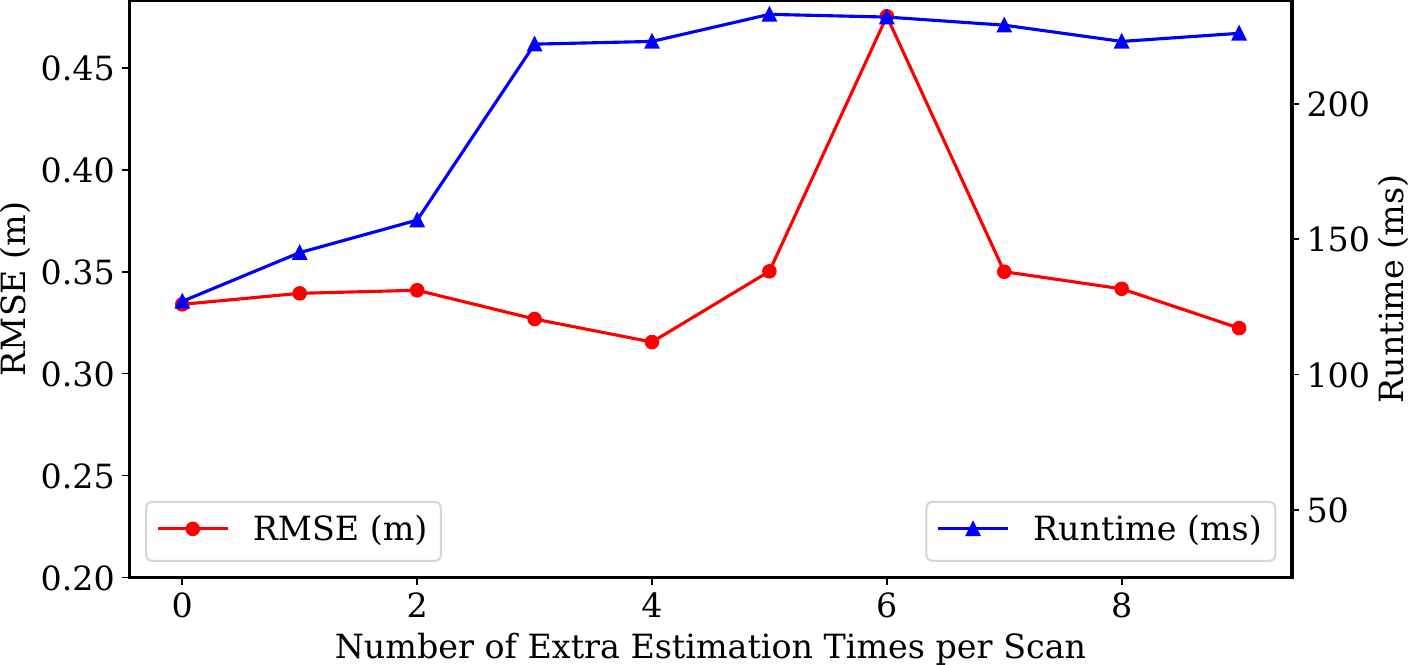}
 	\caption{Root mean squared error vs. the number of estimation times per lidar scan for STEAM-LO for sequence 01-Short of the Newer College Dataset.}
 	\label{fig:num_extra_states_lo}
 \end{figure}

 \begin{figure}[t]
	\centering
	\includegraphics[width=0.8\columnwidth]{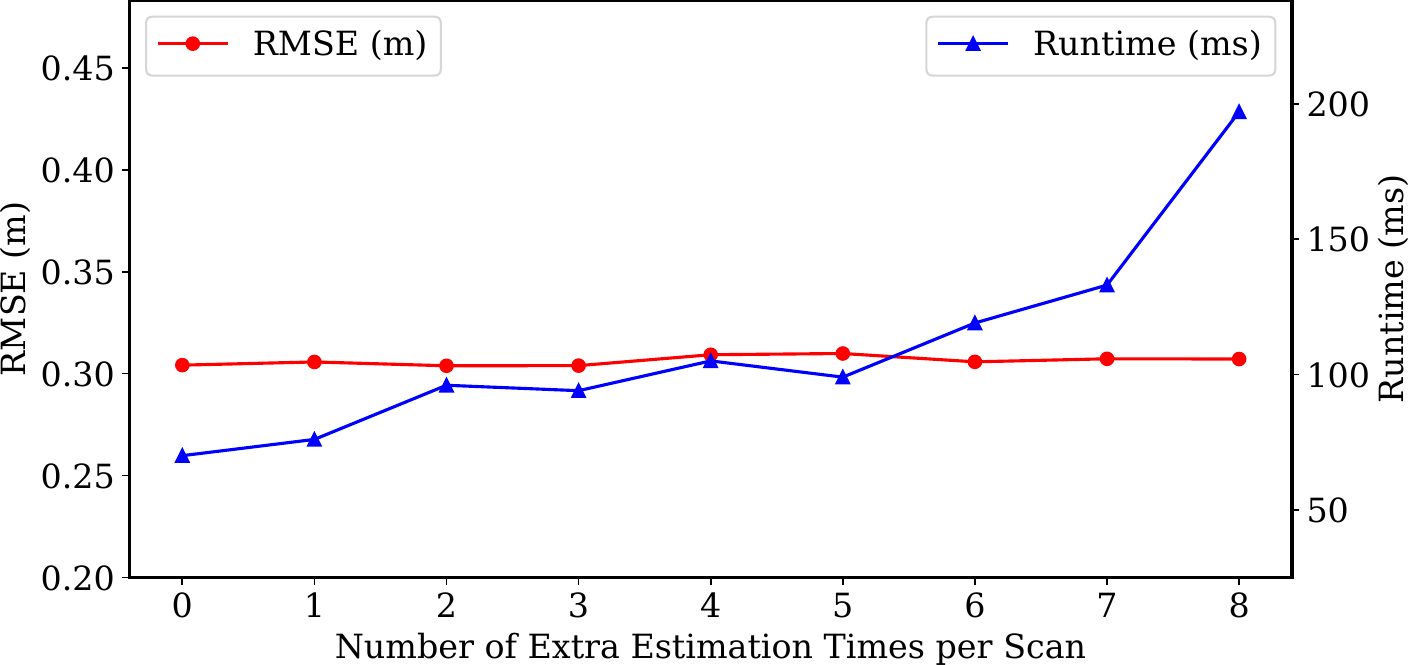}
	\caption{Root mean squared error vs. the number of estimation times per lidar scan for STEAM-LIO for sequence 01-Short of the Newer College Dataset.}
	\label{fig:num_extra_states_lio}
\end{figure}

Here, we analyze the effect of varying the number of additional evenly spaced estimation times for each lidar scan where the default, zero, refers to having an estimation time at the beginning and at the end of each scan. As shown in Figure~\ref{fig:num_extra_states_lo}, STEAM-LO's performance improves slightly when increasing the number of extra estimation times to four. However, the improvement does not continue when increasing the number of estimation times further. In addition, the runtime increases substantially by doing so. For STEAM-LIO, the performance is relatively flat when increasing the number of estimation times as can be seen in Figure~\ref{fig:num_extra_states_lio}. Similar experiments varying the number of estimation times or basis functions were previously presented in \cite{furgale_ijrr15, johnson_arxiv24}.

We also analyze the effect of downsampling the number of unique lidar timestamps. The number of points is unchanged, but the associated timestamps are rounded so that the effective timestamp frequency is reduced. This has the effect of requiring less continuous-time interpolations of the state and also less Jacobians need to be computed during optimization. Overall, the effect is that the continuous-time deskewing is made more coarse and the required runtime is reduced.

In Figure~\ref{fig:lidar_frequency_lo}, STEAM-LO's performance worsens noticeably as the lidar timestamp frequency is reduced. However, at lower timestamp frequencies, STEAM-LO becomes real-time. For STEAM-LIO, the performance worsens only modestly as the timestamp frequency is reduced and the reduction in runtime is also more modest as depicted in Figure~\ref{fig:lidar_frequency_lio}.

 \begin{figure}[t]
	\centering
	\includegraphics[width=0.8\columnwidth]{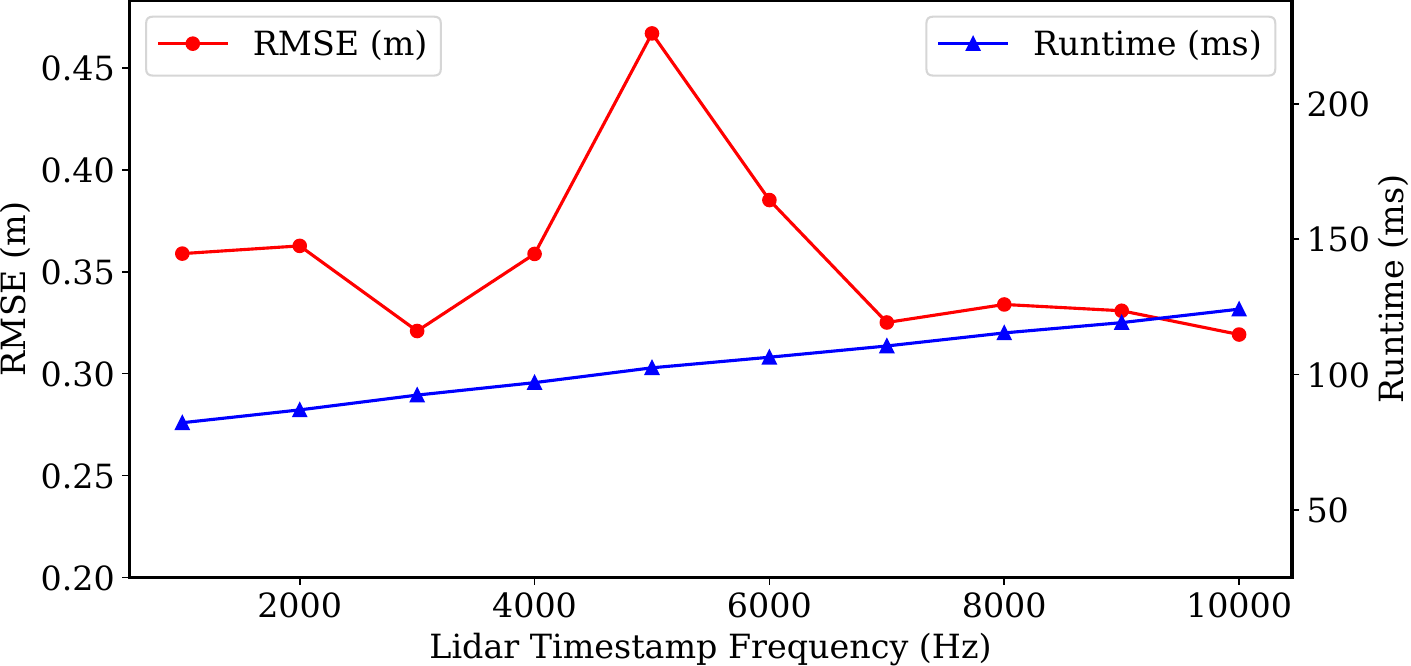}
	\caption{Root mean squared error (RMSE) vs. the lidar timestamp frequency for STEAM-LO for sequence 01-Short of the Newer College Dataset.}
	\label{fig:lidar_frequency_lo}
\end{figure}

\begin{figure}[t]
	\centering
	\includegraphics[width=0.8\columnwidth]{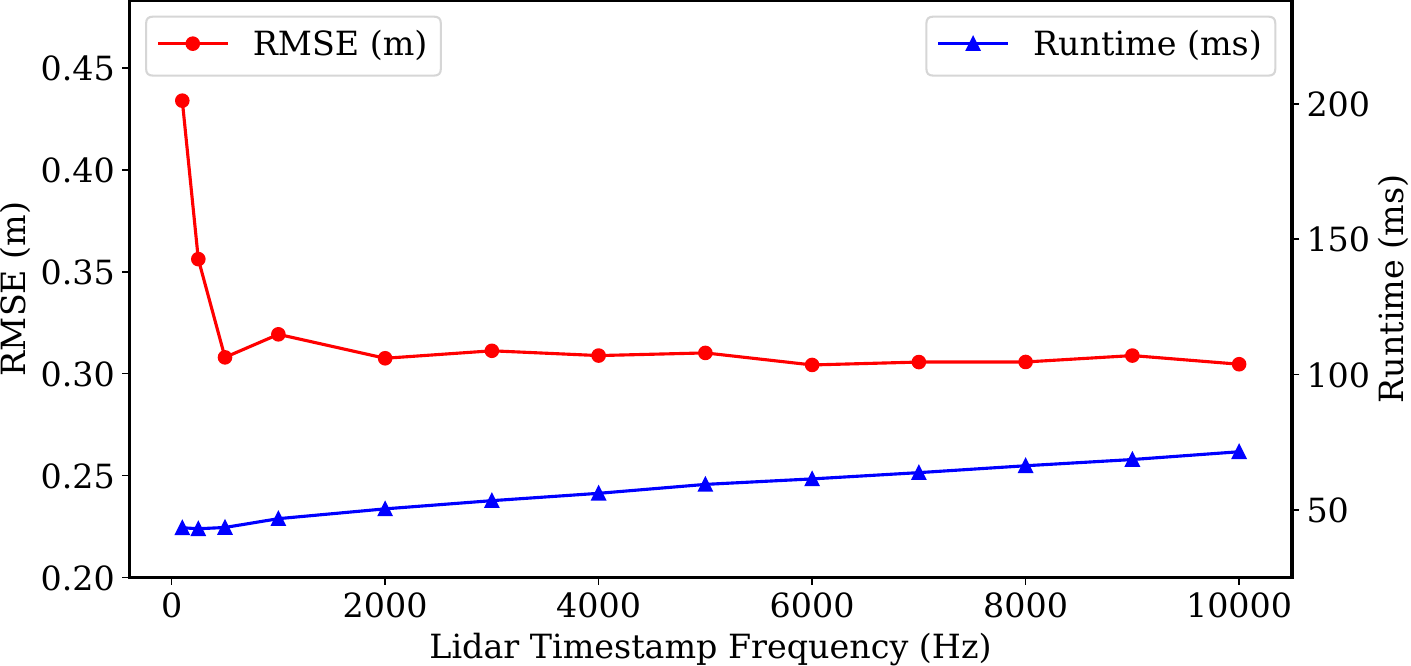}
	\caption{Root mean squared error (RMSE) vs. the lidar timestamp frequency for STEAM-LIO for sequence 01-Short of the Newer College Dataset.}
	\label{fig:lidar_frequency_lio}
\end{figure}

Finally, we analyze the effect of varying the parameters of the diagonal of $\bm{Q}$, the power spectral density matrix, where the default parameters are $\text{diag}(\bm{Q}) = \{50,50,50,5,5,5\}$.

\begin{table}[b]
	\centering
	\footnotesize
	\caption{ATE results (m) on sequence 01-Short of the Newer College Dataset when varying $\text{diag}(\bm{Q}) = \{50,50,50,5,5,5\}$.}
	\begin{tabular}{l ? c c}
		\toprule
		\midrule
		$\bm{Q}$     & STEAM-LO & STEAM-LIO \\
		\midrule
		$\times 1/4$ & 0.3098   & 0.3080    \\
		$\times 1/2$ & 0.3287   & 0.3071    \\
		$\times 1$   & 0.3398   & 0.3042    \\
		$\times 2$   & Failed   & 0.3057    \\
		$\times 4$   & Failed   & 0.3083    \\
		$\times 8$   & Failed   & 0.3056   \\
		\bottomrule
	\end{tabular}
	\label{tab:newer_college_varying_qc}
\end{table}

In Table~\ref{tab:newer_college_varying_qc}, we can see that STEAM-LO is more sensitive to varying the parameters of $\bm{Q}$ than STEAM-LIO. The performance of STEAM-LO improves substantially when decreasing $\bm{Q}$. The effect of this reduction is to increase the weight of penalizing the state estimates from deviating from a constant velocity. However, when increasing the parameters of $\bm{Q}$, STEAM-LO fails. 


}

\subsection{Boreas Results}

\begin{figure}[t]
	\centering
	
	\begin{tikzpicture} [
		arrow/.style={>=latex,red, line width=1.25pt},
		block/.style={rectangle, draw, fill=white, fill opacity=0.8, minimum width=4em, text centered, rounded corners, minimum height=1.25em, line width=1.25pt, inner sep=2.5pt}]
		\node[inner sep=0pt] (boreas) {\includegraphics[trim=30 0 0 0, clip, width=\columnwidth]{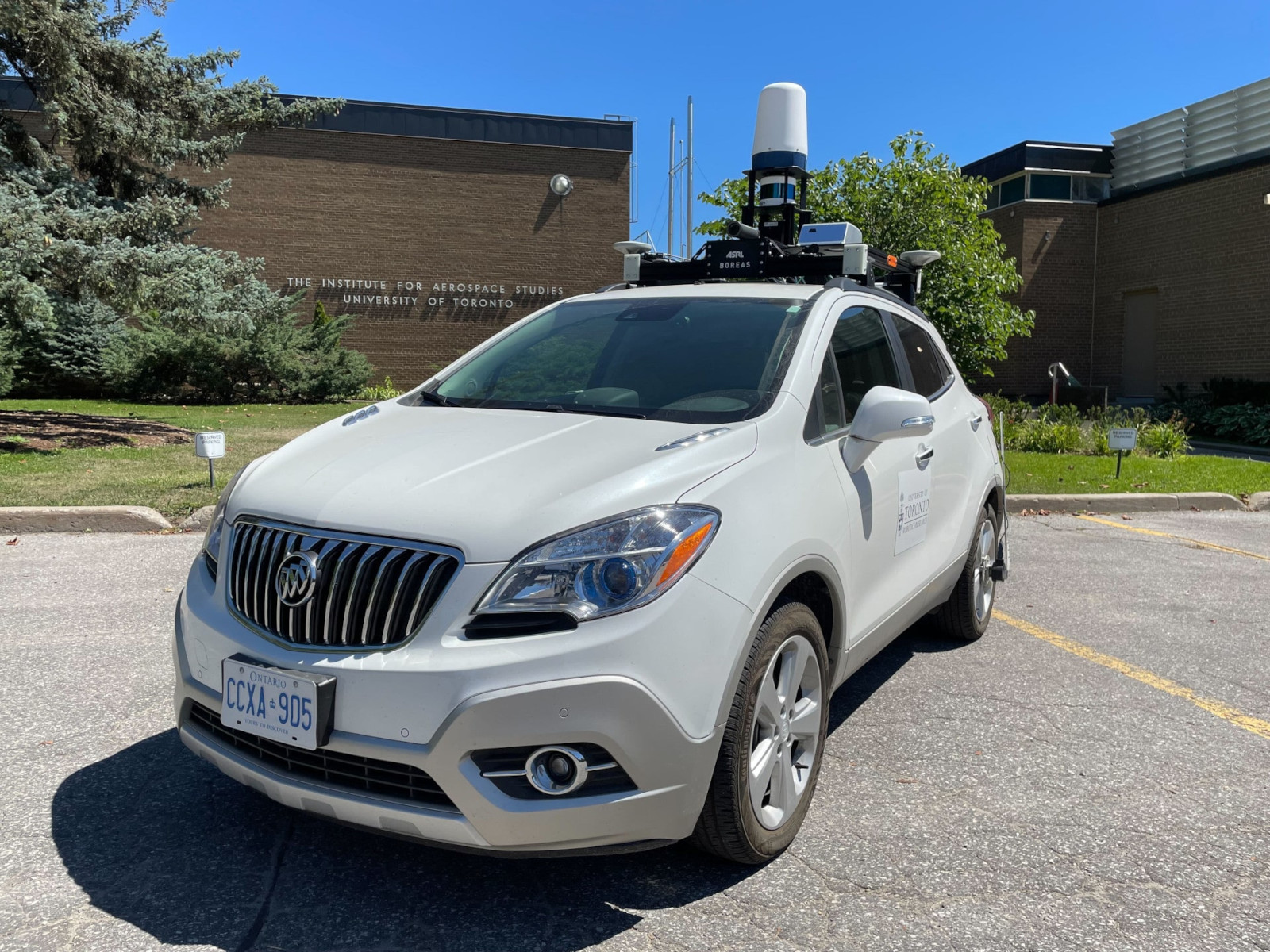}};
		\node (A) at (boreas.center) {};
		\def \L{0.75};

		\node[inner sep=1pt, draw=red, rounded corners, line width=1pt, fill=white, fill opacity=0.8, text opacity=1.0] (aeva) at ($(boreas.center) + (20.5mm, -18mm)$) {\includegraphics[trim=0 0 0 0, clip, width=0.5\columnwidth]{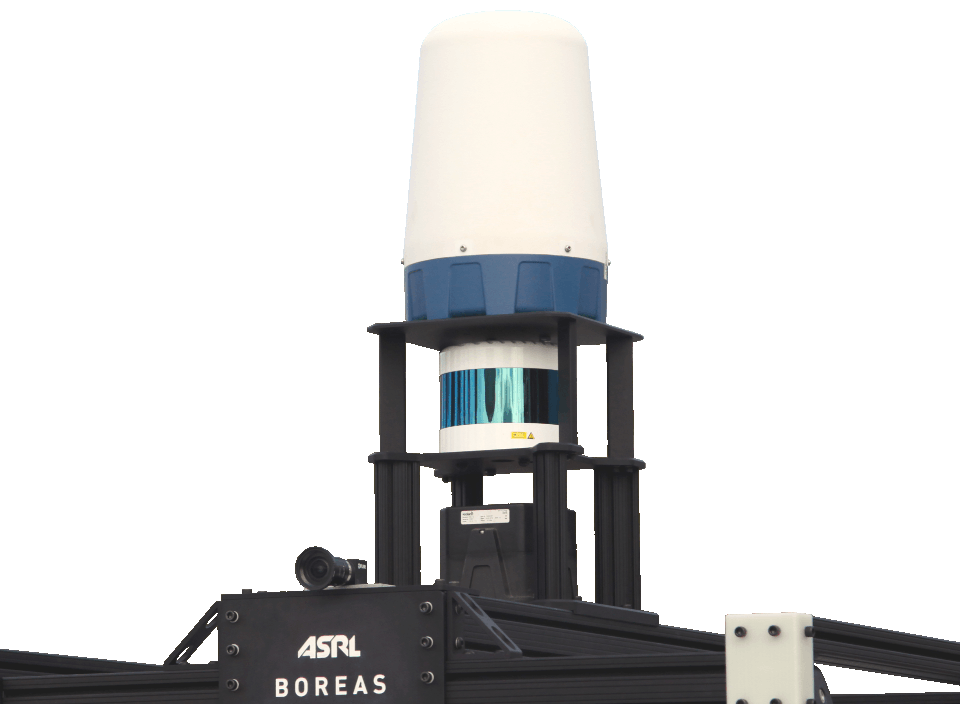}};
		
		\node [block, text opacity=1.0] (radar) at ($ (aeva) + (-14mm, 9mm) $) {\footnotesize\textbf{Radar}};
		\draw[->, arrow] ($ (radar.east) + (0mm, 0mm) $) -- ($ (radar.east) + (4mm, 0mm) $) {};
		
		\node [block, text opacity=1.0] (lidar) at ($ (aeva) + (-14mm, -2mm) $) {\footnotesize\textbf{Lidar}};
		\draw[->, arrow] ($ (lidar.east) + (0mm, 0mm) $) -- ($ (lidar.east) + (4mm, 0mm) $) {};
		
		\node [block, text opacity=1.0] (imu) at ($ (aeva) + (14.5mm, -9.5mm) $) {\footnotesize\textbf{IMU}};
		\draw[->, arrow] ($ (imu.west) + (0mm, 0mm) $) -- ($ (imu.west) + (-3mm, 0mm) $) {};
		
		
	\end{tikzpicture}

	\caption{This figure depicts our data collection platform Boreas which includes a Velodyne Alpha-Prime 128-beam lidar, a Navtech CIR304-H radar, an Applanix GNSS/IMU, and a FLIR Blackfly S camera.}
	\label{fig:boreas}
\end{figure}
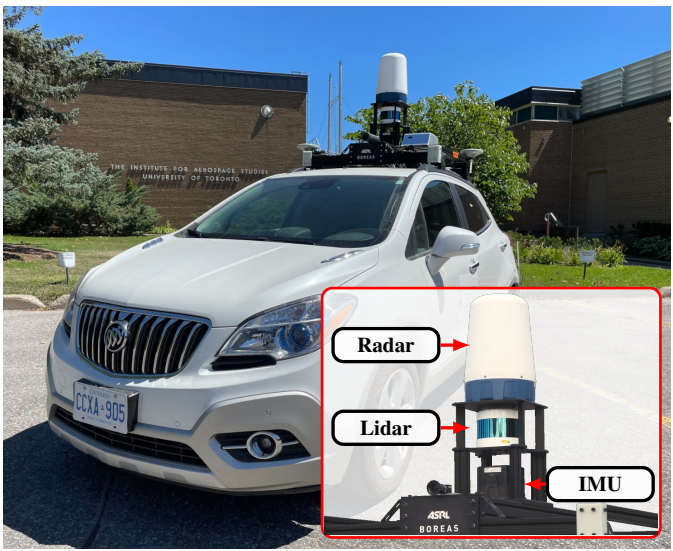

The Boreas dataset was collected at the University of Toronto by driving a repeated route over the course of one year. The dataset features varying seasonal and weather conditions. The sensor suite, depicted in Figure~\ref{fig:boreas} includes a Navtech CIR304-H radar, a 128-beam Velodyne lidar, and an Applanix GNSS/INS. Ground-truth poses were obtained by post-processing all GPS, IMU, and wheel encoder measurements as well as using a subscription for GPS corrections. In total, the test set features 102km or 4.3h of driving data. We extract 200Hz raw IMU measurements from the Applanix logs and take care to ensure that the IMU measurements are not \change{bias-corrected} in any way by the GPS. Table~\ref{tab:boreas_quantitative} shows our quantitative results for this experiment where we compare several variations of our approach: lidar odometry, lidar-inertial odometry, radar odometry, and radar-inertial odometry. We also compare against our previously published work Visual Teach \& Repeat 3 (VTR3) \cite{burnett_ral21}.

It is somewhat surprising to see that our lidar-inertial odometry does not do much better than our lidar odometry here. Our hypothesis is that for relatively slow moving ground vehicles as in the Boreas dataset, our continuous-time lidar odometry is \change{sufficient to compensate} for the motion distortion in the pointcloud. As such, the additional inertial inputs do not significantly improve performance. On the other hand, we can see that for radar odometry, including an IMU results in a significant improvement of \change{43\%}. Our interpretation of this result is that, due the sparsity and noisiness of the radar data, there is more room for improvement by including an IMU. \change{We improve our results even further for our competition submission, STEAM-RIO++, described in the appendix.}

Another \change{observation} is that, when we evaluate our lidar odometry in $SE(2)$, we observe a significant gap in the performance of lidar and radar odometry. This is somewhat contrary to what has been shown in prior work where radar odometry appeared to be getting close to the performance of lidar \cite{adolfsson_tro22}. One important caveat here is that the underlying ground truth is in $SE(3)$ whereas radar odometry is being estimated in $SE(2)$. As such, we have to project the ground truth from 3D to 2D before comparing it to the radar odometry estimates. This projection becomes less accurate as the trajectory length increases. The KITTI odometry metric computes the average drift over all subsequences of lengths $\{100m, 200m, \cdots, 800m\}$. Thus, it is likely that a large part of this apparent radar odometry error is due to this projection error. It appears that we need an improved set of metrics to better compare radar and lidar odometry in a fair manner. We leave this as an area of future work.

\begin{figure}[t]
	\centering
	\includegraphics[width=1.0\columnwidth]{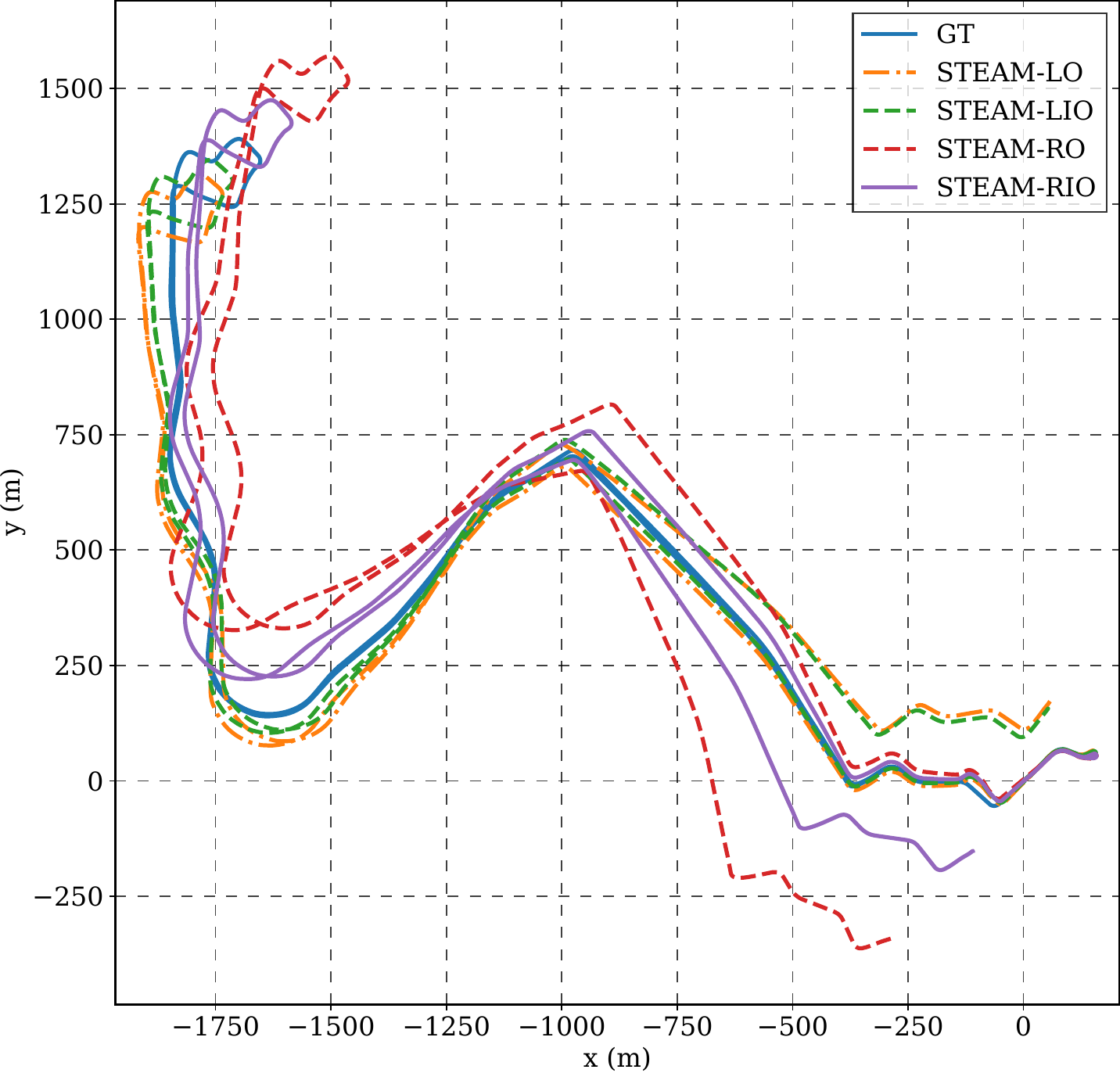}
	\caption{\change{In this figure we compare the performance of odometry approaches presented in this work using the Boreas dataset. The depicted sequence is 2021-01-26-10-59 which was collected during a snowstorm.}
	}
	\label{fig:boreas_odom}
\end{figure}

\begin{figure}[t]
	\centering
	\subfigure[Lidar Odometry Drift vs. Path Length]{\includegraphics[width=1.0\columnwidth]{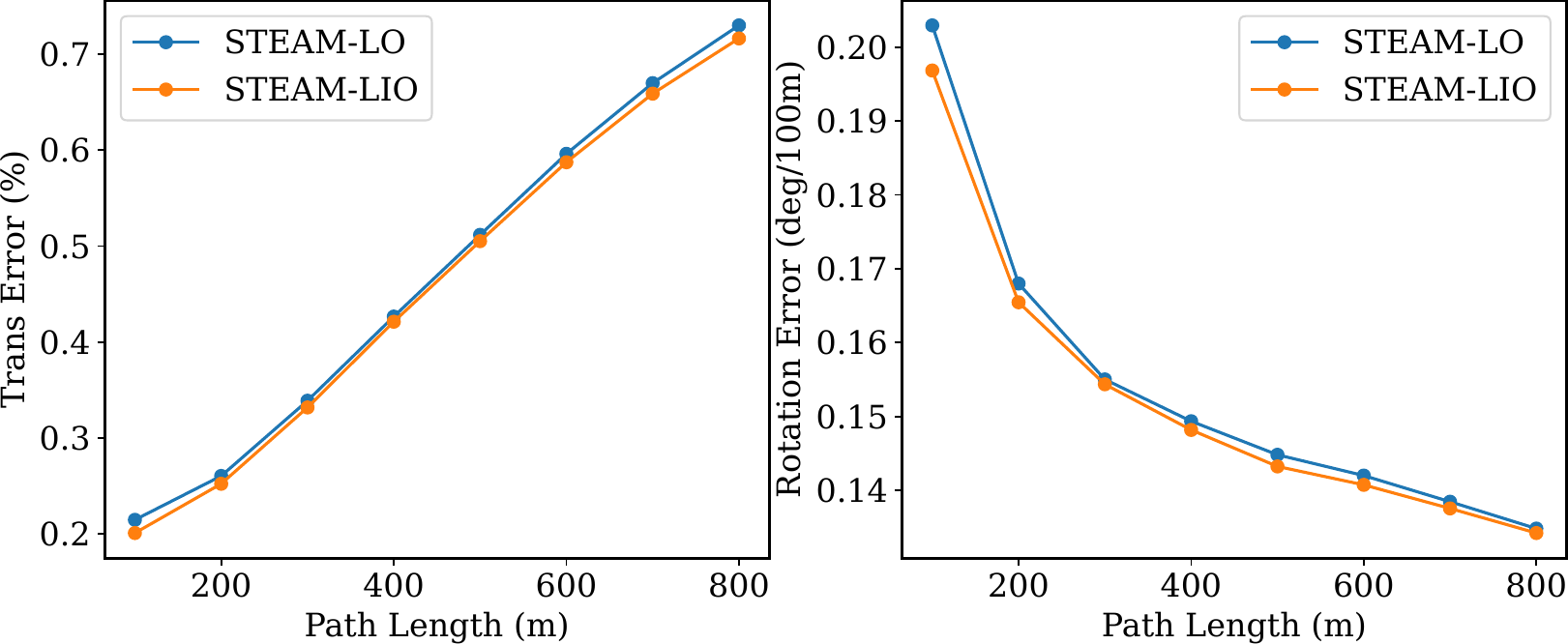}}
	\subfigure[Radar Odometry Drift vs. Path Length]{\includegraphics[width=1.0\columnwidth]{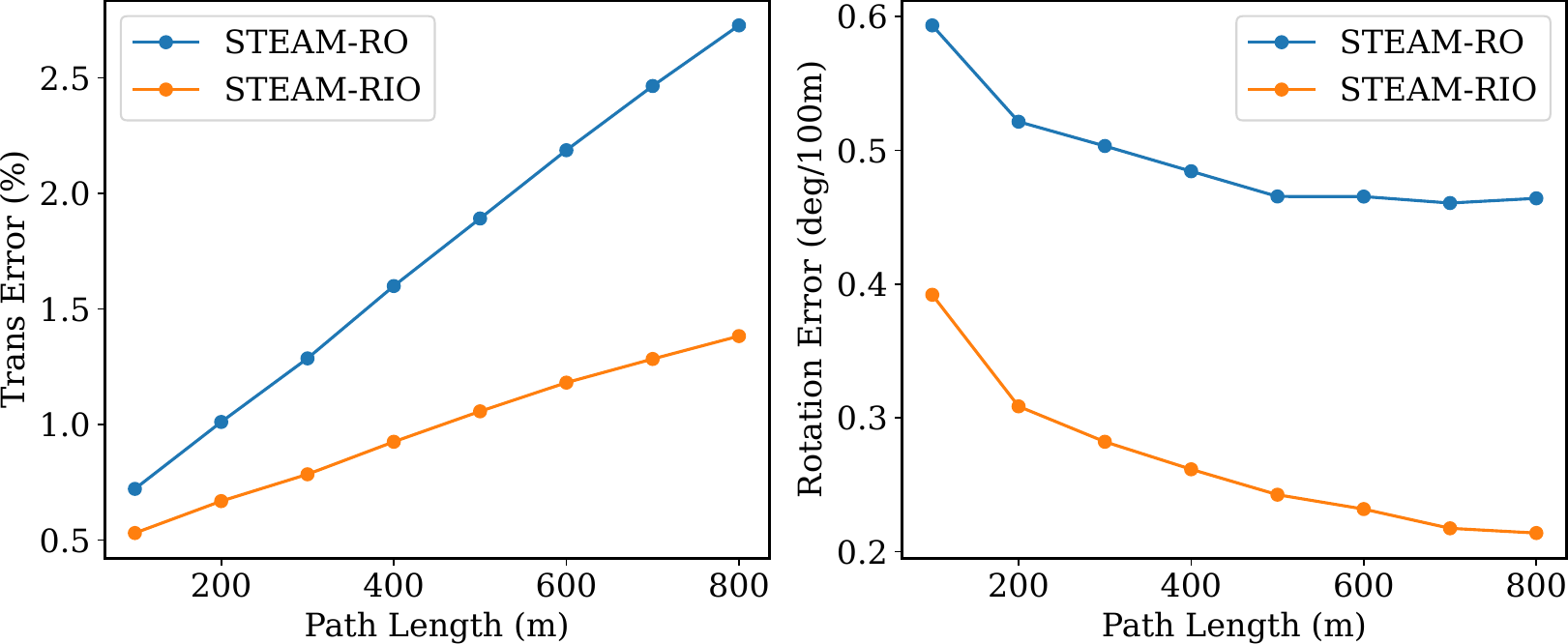}}
	\caption{\change{Here we compare odometric drift vs. path length for lidar and radar odometry. Lidar-inertial odometry performs similarly to lidar odometry. However, radar-inertial odometry improves noticeably over radar odometry.}}
	\label{fig:boreas_kitti-errs}
\end{figure}

\begin{figure}[H]
	\centering
	\includegraphics[width=1.0\columnwidth]{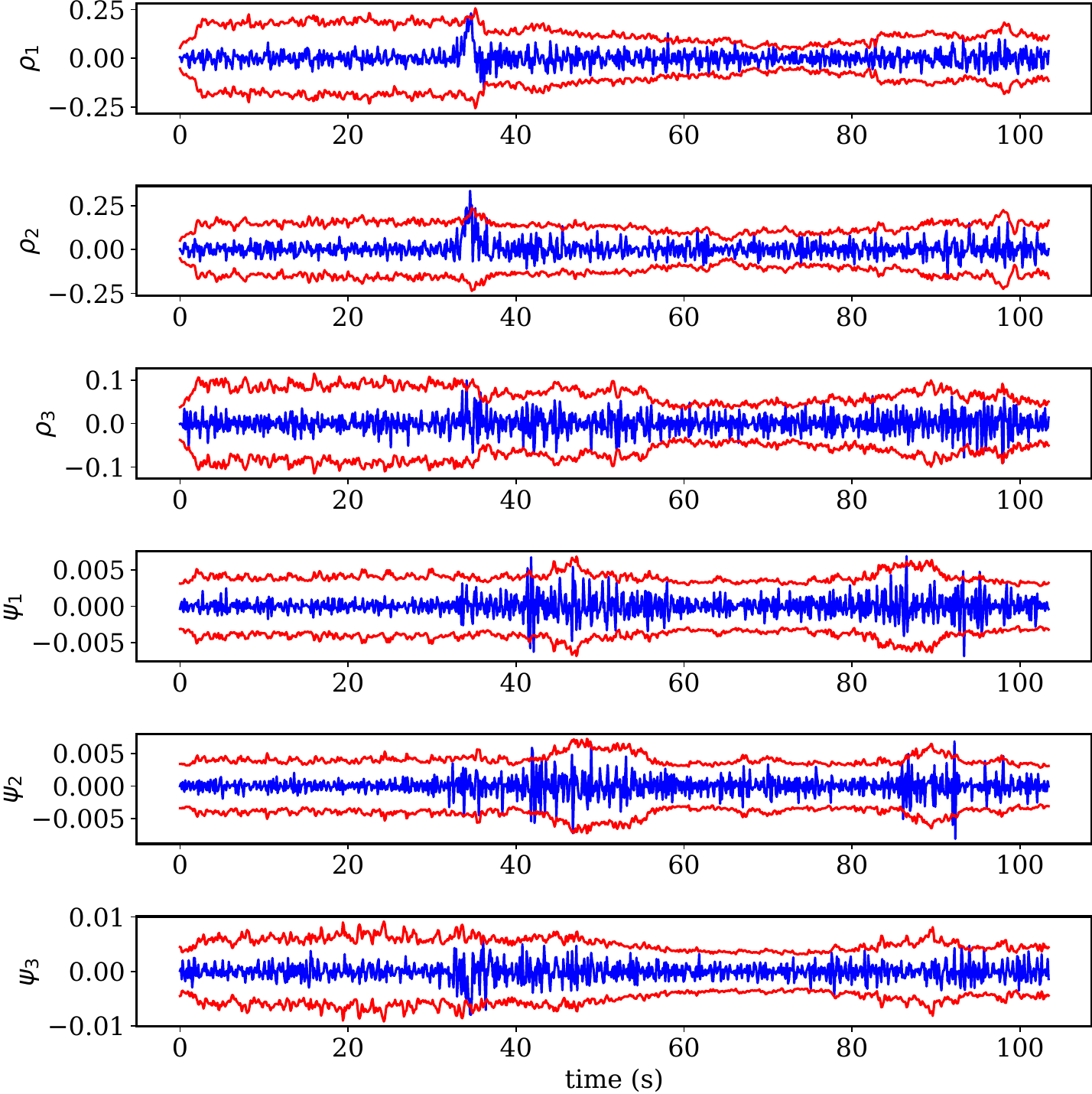}
	\caption{This figure plots error vs. time for our lidar-inertial odometry when compared to ground truth on the first 100 seconds of sequence 2021-01-26-10-59. The red lines denote the estimated 3$\sigma$ uncertainty bounds. Each row represents a dimension of the log map of the pose error where $\rho$ is a translational dimension and $\psi$ is a rotational dimension.}
	\label{fig:boreas_odom_errs}
\end{figure}

Figure~\ref{fig:boreas_odom} shows a qualitative example of trajectories estimated by our approach on the Boreas dataset. We observe that our lidar odometry remains close to the ground truth while our lidar-inertial odometry achieves a similar result. It can also be observed that our radar-inertial odometry is notably better than our radar-only odometry. We also compare the odometry metrics as a function of path length in Figure~\ref{fig:boreas_kitti-errs} where we observe that including an IMU results in only a minor improvement for lidar odometry but results in a significant improvement for radar odometry. In Figure~\ref{fig:boreas_odom_errs}, we plot the odometry errors vs. time where we compare frame-to-frame odometry estimates vs. the ground truth. We also plot the estimated $3\sigma$ uncertainty bounds in red. Note that our approach is quite consistent, our estimated uncertainty does a good job of capturing the actual spread of the error. For each new lidar frame, our approach estimates the pose of the vehicle in a drifting map frame $\hat{\mathbf{T}}_k = \hat{\mathbf{T}}_{v,i}(t_k)$ where $t_k$ corresponds to the temporal middle of the scan. We first compose two of these estimates to obtain a relative odometry pose change,
\begin{equation}
	\hat{\mathbf{T}}_{k,k-1} = \hat{\mathbf{T}}_k \hat{\mathbf{T}}_{k-1}^{-1}.
\end{equation}

 The error that we compute is

\begin{equation}
	\boldsymbol{\xi}_{k,k-1} = \ln\left( \hat{\mathbf{T}}_{k,k-1}\mathbf{T}_{k,k-1}^{-1} \right)^\vee, \label{eq:odom_Error}
\end{equation} where $\mathbf{T}_{k,k-1}$ is the ground-truth odometry pose. We estimate a covariance $\hat{\mathbf{P}}_{k,k}$ for the pose of each lidar frame by interpolating the covariance over the sliding window at time $t_k$. See \cite[\textsection 11.3.2]{barfoot_ser17} for details. The covariance of $\hat{\mathbf{T}}_{k,k-1}$, $\text{cov}(\hat{\mathbf{T}}_{k,k-1}) = \boldsymbol{\Sigma}_{k,k-1}$,  is obtained using  \cite{barfoot_ser17}

\begin{equation}
	\boldsymbol{\Sigma}_{k,k-1} \approx \hat{\mathbf{P}}_{k,k} + \T_{k,k-1} \hat{\mathbf{P}}_{k-1,k-1} \T_{k,k-1}^T, \label{eq:cov_approx}
\end{equation} where $\T_{k,k-1} = \text{Ad}(\hat{\mathbf{T}}_{k,k-1})$. \ref{eq:odom_Error} and \ref{eq:cov_approx} are then used to produce the plots in Figure~\ref{fig:boreas_odom_errs}. We compute the normalized estimation error squared (NEES) using

\begin{equation}
	\text{NEES} = \sum_{k=1}^K \frac{\boldsymbol{\xi}_{k,k-1}^T \boldsymbol{\Sigma}_{k,k-1}^{-1} \boldsymbol{\xi}_{k,k-1}}{\text{dim}(\boldsymbol{\xi}_{k,k-1})K}.
\end{equation} For the lidar-inertial odometry shown in Figure~\ref{fig:boreas_odom} and Figure~\ref{fig:boreas_odom_errs}, we obtain a NEES of 1.04 where an ideal value is 1.0. This means that our estimator is slightly overconfident here.

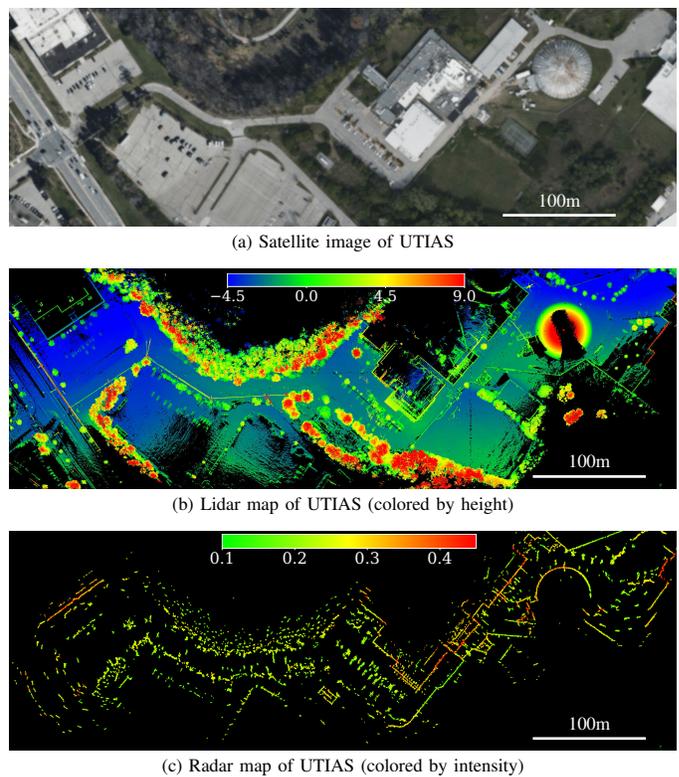
\begin{figure}[H]
	\centering
		\begin{tikzpicture}
			\node[inner sep=0pt] (img) {\includegraphics[trim=0 50 0 50, clip,width=1.0\columnwidth]{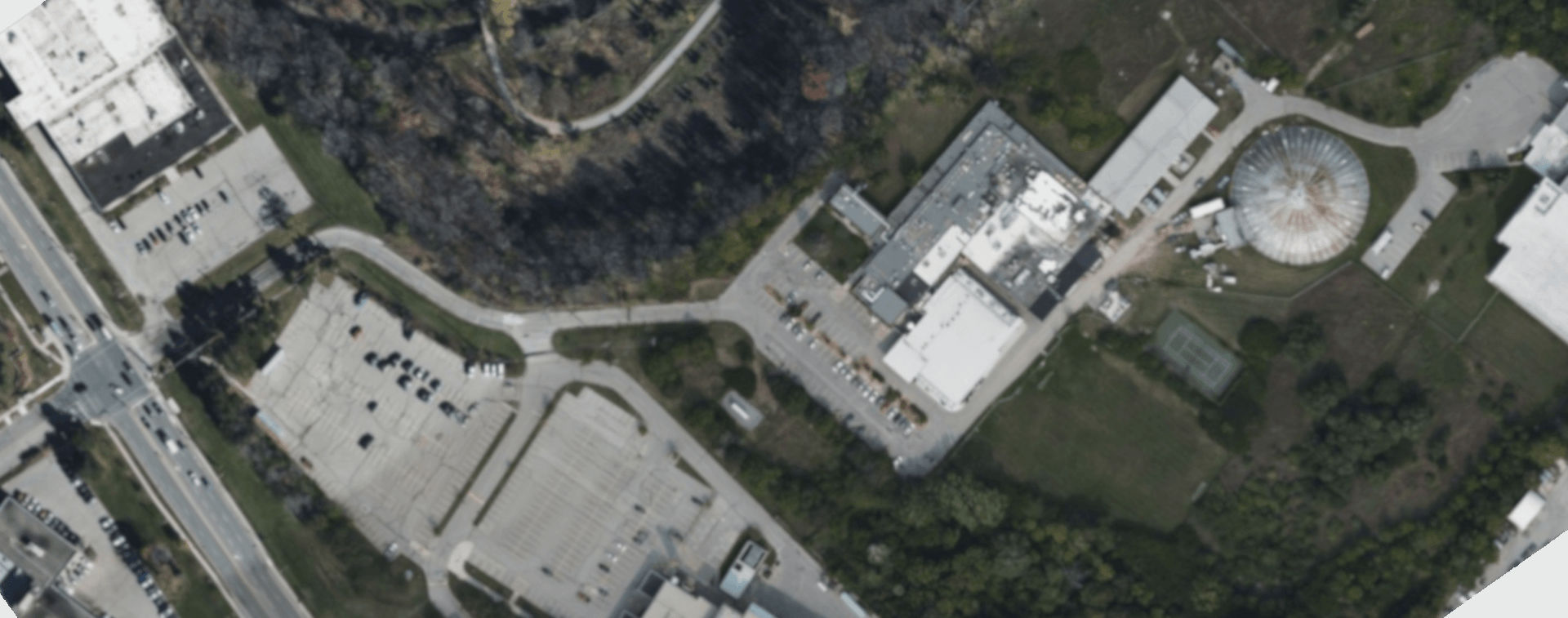}};
			\node[inner sep=0pt, below=of img, yshift=9mm] {\scriptsize (a) Satellite image of UTIAS};
			\node[inner sep=0pt, below=of img, yshift=4.5mm] (img2)  {\includegraphics[trim=0 50 0 50, clip,width=1.0\columnwidth]{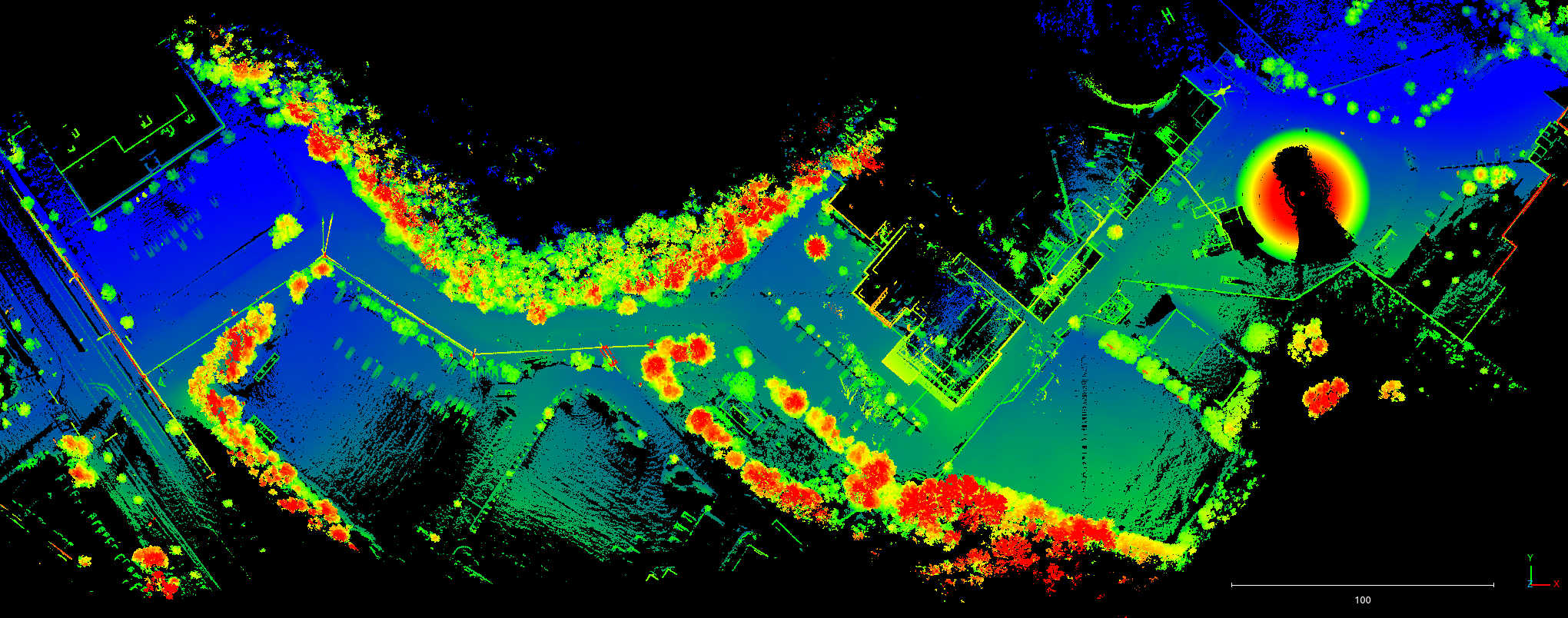}};
			\node[inner sep=0pt, below=of img2, yshift=9mm] {\scriptsize (b) Lidar map of UTIAS (colored by height)};
			\node[inner sep=0pt, below=of img2, yshift=4.5mm] (img3)  {\includegraphics[trim=0 50 0 50, clip,width=1.0\columnwidth]{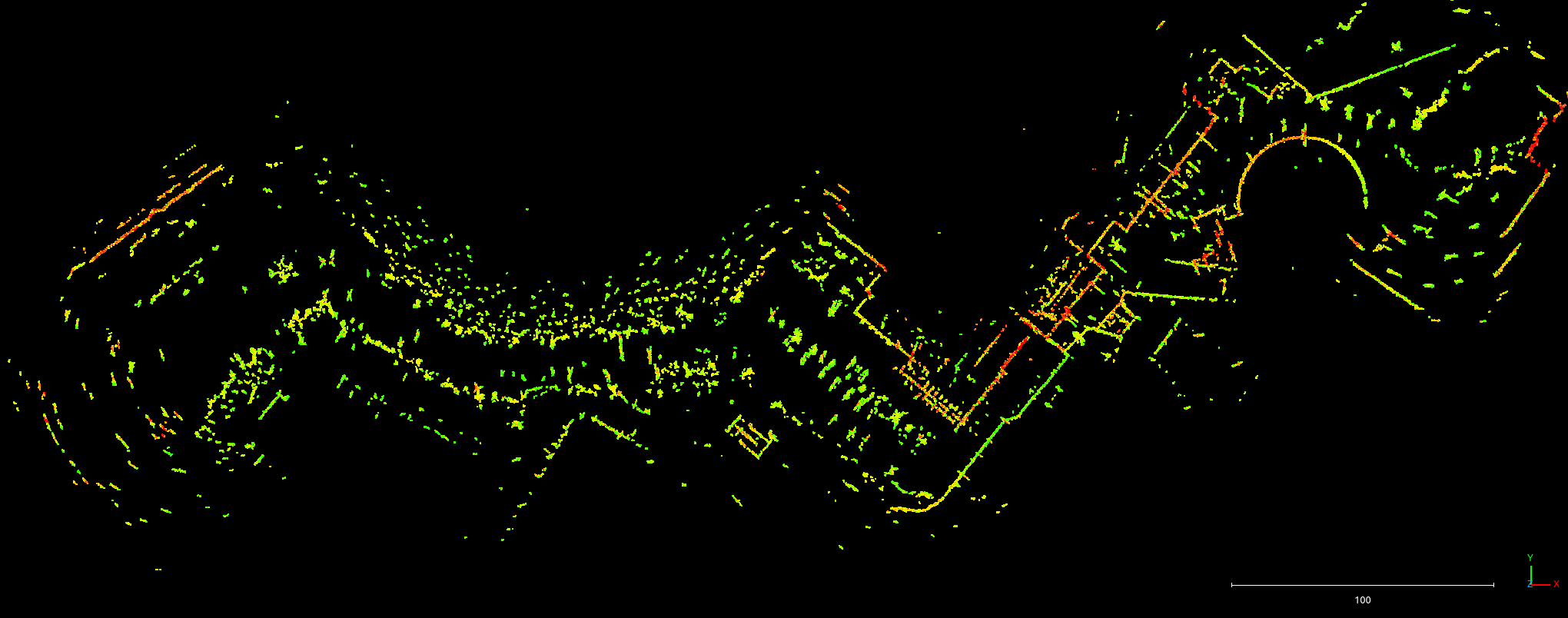}};
			\node[inner sep=0pt, below=of img3, yshift=9mm] {\scriptsize (c) Radar map of UTIAS (colored by intensity)};
			
			\node (p0) [right of=img, xshift=10.0mm, yshift=-13.mm] {};
			\node (p1) [right of=img, xshift=27.5mm, yshift=-13.mm] {};
			\draw[white, line width=1pt] (p0) -- (p1);
			\node[white] (legend) at ($(p0)!0.5!(p1) + (0mm, 2mm)$) {{\scriptsize 100m}};
			
			\node[inner sep=0pt] (cbar1) [above of=img2, yshift=2mm] {\includegraphics[width=0.4\columnwidth]{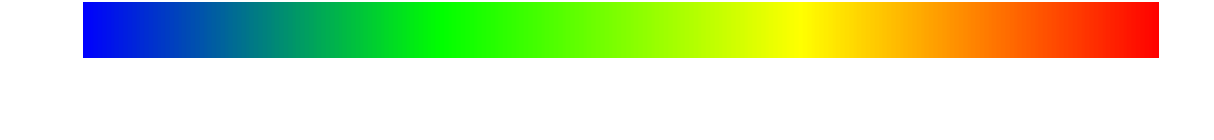}};
			\node (p0) [right of=img2, xshift=14.mm, yshift=-13.mm] {};
			\node (p1) [right of=img2, xshift=31.5mm, yshift=-13.mm] {};
			\draw[white, line width=1pt] (p0) -- (p1);
			\node[white] (legend) at ($(p0)!0.5!(p1) + (0mm, 2mm)$) {{\scriptsize 100m}};
			
			\node[inner sep=0pt] (cbar2) [above of=img3, yshift=2mm] {\includegraphics[width=0.4\columnwidth]{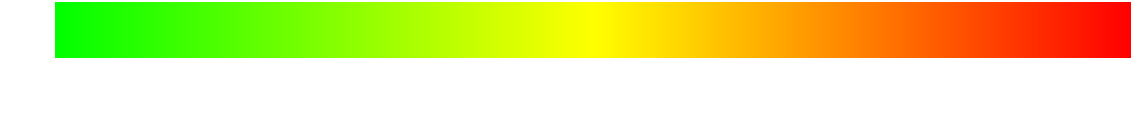}};
			\node (p0) [right of=img3, xshift=14mm, yshift=-13.mm] {};
			\node (p1) [right of=img3, xshift=31.5mm, yshift=-13.mm] {};
			\draw[white, line width=1pt] (p0) -- (p1);
			\node[white] (legend) at ($(p0)!0.5!(p1) + (0mm, 2mm)$) {{\scriptsize 100m}};
		\end{tikzpicture}
	\caption{This figure displays maps generated of the University of Toronto Institute for Aerospace Studies (UTIAS). The data was obtained using the first 134 seconds of the 2021-02-09 Boreas sequence.}
	\label{fig:maps_utias1}
\end{figure}

In Figure~\ref{fig:maps_utias1}, we compare the maps generated using our lidar-inertial odometry and our radar-inertial odometry. The lidar map is colored by height and is displayed using a top-down orthographic projection. Our lidar-inertial odometry has minimal drift and so the map that it generates aligns quite well with satellite imagery.

It is clear that a high level of detail is being captured by the lidar map. In order to produce the radar map \change{in the figure}, we took snapshots of the \change{online} radar map every 10m and aligned these submaps using the estimated odometry. We then removed noisy detections from the radar map by performing a radius outlier removal of points with less than two neighbors within a radius of 0.25m and a statistical outlier removal of points greater than one standard deviation above the average point-to-point distance. The radar map drifts slightly with respect to the lidar map. Note that we did not use any loop closures to generate these maps so there is room for improvement. Even though these maps contain some drift, we showed in our prior work that maps only need to be locally consistent to enable accurate localization \cite{burnett_ral22}.






\section{Conclusions}

In this work, we showed that our Gaussian process motion prior is often sufficient to \change{compensate for} motion-distorted lidar data \change{when there is sufficient geometric features}. However, \change{in challenging scenarios such as in the Newer College Dataset}, we showed that our continuous-time lidar odometry could be augmented with \change{IMU} measurements to handle these conditions. Even in the presence of aggressive motion, the majority of the improvement resulted from the inclusion of gyroscope measurements whereas the addition of accelerometer measurements yielded only a minor additional improvement. We showed that we could improve our radar odometry by \change{43\%} by including inertial measurements. Contrary to previous work, we showed that there is still a significant gap between the performance of radar and lidar odometry under nominal conditions. Part of this gap may be explained by difficulties comparing 3D and 2D odometry estimates. Improved metrics for this purpose is an area of future work. Including body-centric acceleration in the state is another area of future work.

\section*{Acknowledgments}
We thank General Motors for their donation of the Buick and Applanix for their support in post-processing the ground truth for the Boreas dataset and providing tools to extract raw IMU data. We also thank the Ontario Research Fund for supporting this work.





\appendix

\subsection{Interpolation Jacobians}

In this work, we build continuous-time measurement factors by making use of the posterior Gaussian process interpolation formula. In order to do this, we need to compute the Jacobians of the perturbation to the interpolated state $\delta \mathbf{x}(\tau)$ with respect to the state perturbations at the bracketing estimation times $\delta \mathbf{x}_k, \delta \mathbf{x}_{k+1}$. Perturbations to the state at estimation times are defined as
\begin{subequations}
\begin{align}
	\mathbf{T}_k &= \exp(\e_k^\wedge) \mathbf{T}_{\text{op},k},\\
	\w_k &= \w_{\text{op},k} + \boldsymbol{\eta}_k.
\end{align}
\end{subequations}

In order to compute the interpolation Jacobians, we first need to linearize some expressions contained in \eqref{eq:interp}. When we evaluate the local Markovian variable at the endpoints of the local GP, we get the following results,
\begin{subequations}
\begin{align}
	\hat{\boldsymbol{\gamma}}_k(t_k) &= \begin{bmatrix} \mathbf{0} \\ \hat{\boldsymbol{\varpi}}_k \end{bmatrix}, \\ \hat{\boldsymbol{\gamma}}_k(t_{k+1}) &= \begin{bmatrix} \ln\left( \mathbf{\hat{T}}_{k+1} \mathbf{\hat{T}}_{k}^{-1} \right)^\vee \\ \J \left( \ln\left( \mathbf{\hat{T}}_{k+1} \mathbf{\hat{T}}_{k}^{-1} \right)^\vee \right)^{-1} \hat{\boldsymbol{\varpi}}_{k+1} \end{bmatrix}. \label{eq:gamma2}
\end{align}
\end{subequations} Next, we linearize
\begin{align}
	\ln\left( \mathbf{T}_{k+1} \mathbf{T}_{k}^{-1} \right)^\vee &\approx \ln\left( \mathbf{T}_{\text{op}, k+1} \mathbf{T}_{\text{op},k}^{-1} \right)^\vee \\
	&+ \J_{\text{op},k+1,k}^{-1} (\e_{k+1} - \T_{\text{op},k+1,k}\e_k ), \nonumber
\end{align} where we have assumed that $\e_{k+1} - \T_{\text{op},k+1,k}\e_k$ is small and we have defined 
\begin{subequations}
\begin{align}
	\T_{\text{op},k+1,k} &= \T_{\text{op},k+1}\T_{\text{op},k}^{-1},\\ \J_{\text{op},k+1,k} &= \J\left(\ln(\T_{\text{op},k+1}\T_{\text{op},k}^{-1})^\vee \right),
\end{align}
\end{subequations} and $\T_{\text{op},k} = \text{Ad}(\mathbf{T}_{\text{op}, k})$. We also make the following linearization
\begin{align}
	\J \Big( \ln\Big( \mathbf{\hat{T}}_{k+1} &\mathbf{\hat{T}}_{k}^{-1} \Big)^\vee \Big)^{-1}  \approx \J_{\text{op},k+1,k}^{-1} \\
	&- \frac{1}{2}\left( \J_{\text{op},k+1,k}^{-1}(\e_{k+1} - \T_{\text{op},k+1, k} \e_k) \right)^\curlywedge, \nonumber
\end{align} where we have again assumed that $\e_{k+1} - \T_{\text{op},k+1,k}\e_k$ is small and we have approximated the inverse left Jacobian with $\J^{-1}(\mathbf{x}) \approx \mathbf{1} - \frac{1}{2}\mathbf{x}^\curlywedge$. The interpolated local variables between estimation times $t_k, t_{k+1}$ are defined as
\begin{subequations}
\begin{align}
	\boldsymbol{\xi}_k(\tau) &= \boldsymbol{\Lambda}_1(\tau) \hat{\boldsymbol{\gamma}}_k(t_k) + \boldsymbol{\Psi}_1(\tau) \hat{\boldsymbol{\gamma}}_k(t_{k+1}), \\
	\dot{\boldsymbol{\xi}}_k(\tau) &= \boldsymbol{\Lambda}_2(\tau) \hat{\boldsymbol{\gamma}}_k(t_k) + \boldsymbol{\Psi}_2(\tau) \hat{\boldsymbol{\gamma}}_k(t_{k+1}).
\end{align}
\end{subequations}

The general formula for obtaining the interpolation Jacobians for perturbations to the pose and body-centric velocity is as follows:
\begin{subequations}
\begin{align}
	\frac{\partial \delta \mathbf{T}(\tau)}{\partial \mathbf{x}} &= \J_{\text{op},\tau, k} \frac{\partial \x_k(\tau)}{\partial \mathbf{x}} + \T_{\text{op},\tau,k} \frac{\partial \e_k}{\partial \mathbf{x}}, \\
	\frac{\partial \delta \w(\tau)}{\partial \mathbf{x}} &= \J_{\text{op},\tau,k} \frac{\partial \dot{\boldsymbol{\xi}}_k(\tau)}{\partial \mathbf{x}}  -\frac{1}{2} \dot{\x}^\curlywedge_{\text{op},\tau}  \frac{\partial \boldsymbol{\xi}_k(\tau)}{\partial \mathbf{x}} ,
\end{align}
\end{subequations} where the Jacobians of the local variable $\x_k(\tau)$ with respect to state perturbations at the bracketing times are given by

\begin{subequations}
\begin{align}
	\frac{\partial \x_k(\tau)}{\partial \e_{k+1}} &= \boldsymbol{\Psi}_{11} \J_{\text{op},k+1,k}^{-1}  + \frac{1}{2} \boldsymbol{\Psi}_{12} \w_{\text{op},k+1}^\curlywedge \J_{\text{op},k+1,k}^{-1}, \\
	\frac{\partial \x_k(\tau)}{\partial \e_k} &= -\left(\frac{\partial \x_k(\tau)}{\partial \e_{k+1}} \right) \T_{\text{op},k+1,k}, \\
	\frac{\partial \x_k(\tau)}{\partial \boldsymbol{\eta}_k} &= \boldsymbol{\Lambda}_{12}, \\
	\frac{\partial \x_k(\tau)}{\partial \boldsymbol{\eta}_{k+1}} &= \boldsymbol{\Psi}_{12} \J_{\text{op},k+1,k}^{-1}.
\end{align}
\end{subequations}
The Jacobians of $\dot{\x}_k(\tau)$ have the same form except that we use the second row of the interpolation matrices, $\boldsymbol{\Psi}_{21}$ instead of $\boldsymbol{\Psi}_{11}$, for example.

\change{
	
\subsection{2024 Radar in Robotics Competition}
	
At ICRA 2024 in Yokohama, Japan, we hosted a radar odometry competition as part of the Radar in Robotics workshop. In order to achieve our competition results, we increased the length of the sliding window from two scans to four. We also switched from a Cauchy loss to a Huber loss with a more restrictive threshold to filter out outliers. We increased the weight given to gyroscope measurements. We also incorporated a version of keyframing where new radar frames were not added to the sliding local map unless the vehicle travelled at least one meter. Putting together all of these changes, our competition submission, STEAM-RIO++, achieved 0.62\% translational drift, a 35\% improvement over STEAM-RIO as shown in Table~\ref{tab:boreas_supplementary}. In the competition, we placed third, where the 2nd place submission was CFEAR with 0.61\% drift \cite{adolfsson_boreascompetition24}. Notably, CFEAR did not use an IMU, and their approach is computationally very efficient. In order to achieve their competition results, they greatly increased the length of their sliding window and employed a coarse-to-fine registration approach. The first-place entry was CFEAR++, achieving 0.51\% drift \cite{li_boreascompetition24}. CFEAR++ is based on CFEAR where the authors also used gyroscope angle estimates as a prior as well as semantic segmentation to retain only points that correspond to buildings.

}

\begin{table}[ht]
	\centering
	\footnotesize
	\caption{Boreas Odometry Supplementary Results (102km / 4.3h): translational drift (\%) / rotational  drift (deg/100m).}
	\begin{tabular}{ l ? c c c}
		\toprule
		\midrule
		\textbf{Boreas} &   STEAM-RO & STEAM-RIO & STEAM-RIO++     \\
		\midrule
		2020-12-04    & \change{1.43 / 0.41} & \change{0.93 / 0.26} & \change{0.76 / 0.20} \\
		2021-01-26    &  \change{1.10 / 0.33} & \change{0.61 / 0.18} & \change{0.50 / 0.24} \\
		2021-02-09    &  \change{1.27 / 0.38} & \change{0.63 / 0.20} & \change{0.40 / 0.13} \\
		2021-03-09    &  \change{1.24 / 0.35} & \change{0.71 / 0.19} & \change{0.58 / 0.17} \\
		2020-04-22    & \change{1.48 / 0.41} & \change{0.99 / 0.27} & \change{0.67 / 0.18} \\
		2021-06-29-18    &  \change{1.55 / 0.46} & \change{1.04 / 0.29} & \change{0.66 / 0.19} \\
		2021-06-29-20    &  \change{1.70 / 0.48} & \change{0.96 / 0.26} & \change{0.75 / 0.20} \\
		2021-09-08    &  \change{2.01 / 0.59} & \change{1.22 / 0.35} & \change{0.74 / 0.21} \\
		2021-09-09    &  \change{2.16 / 0.64} & \change{1.19 / 0.33} & \change{0.56 / 0.15} \\
		2021-10-05    &  \change{2.27 / 0.63} & \change{1.01 / 0.28} & \change{0.58 / 0.16} \\
		2021-10-26    &  \change{1.88 / 0.53} & \change{0.97 / 0.27} & \change{0.63 / 0.18} \\
		2021-11-06    &  \change{1.86 / 0.54} & \change{1.07 / 0.29} & \change{0.74 / 0.21} \\
		2021-11-28    &  \change{1.95 / 0.57} & \change{1.04 / 0.29} & \change{0.56 / 0.16} \\
		\midrule
		\color{blue}{\textbf{Seq. Avg.}} &  \color{blue}{1.68 / 0.49} & \color{blue}{0.95 / 0.27} & \color{blue}{0.62 / 0.18}\\
		\midrule
		$\Delta T$   & \change{115ms} & \change{139ms} & \change{153ms}\\
		\bottomrule
	\end{tabular}
	\label{tab:boreas_supplementary}
\end{table}

\bibliography{IEEEabrv,bib/references}

\newpage

\begin{IEEEbiography}[{\includegraphics[width=1in,height=1.25in,clip,keepaspectratio]{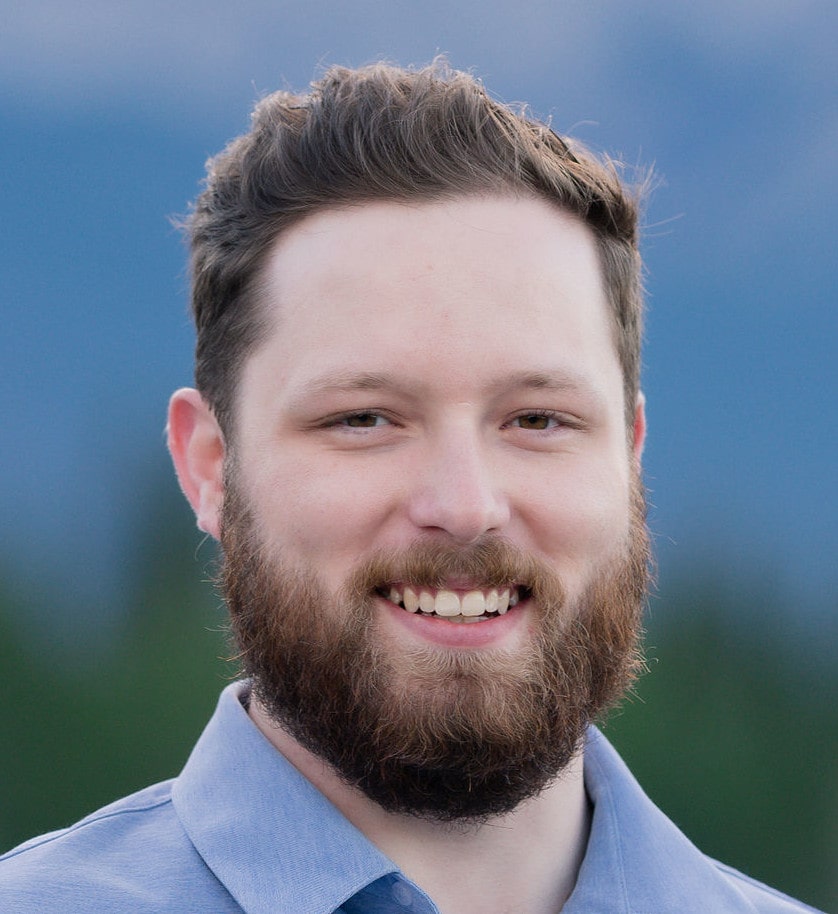}}]{Keenan Burnett} received the B.A.Sc. degree in engineering science from University of Toronto, Toronto, ON, Canada, in 2017 and the M.A.Sc. degree in aerospace engineering from the University of Toronto in 2020. He is a Ph.D. student with the University of Toronto Robotics Institute. His work to date has centered around autonomous vehicles. His current research focuses on mapping and localization in adverse weather conditions through the application of radar and lidar. He is interested in extending the operational capabilities of robots to new and challenging environments.
\end{IEEEbiography}
\vspace{-5mm}

\begin{IEEEbiography}[{\includegraphics[width=1in,height=1.25in,clip,keepaspectratio]{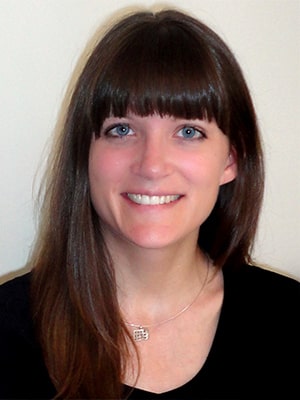}}]{Angela P. Schoellig} received the M.Sc. degree in engineering science and mechanics from the Georgia Institute of Technology, Atlanta, GA, USA, in 2007, the master's degree in engineering cybernetics from the University of Stuttgart, Stuttgart, Germany, in 2008, and the Ph.D. degree in mechanical and process engineering from ETH Zurich, Zurich Switzerland, in 2012. She is currently an Assistant Professor with the Institute for Aerospace Studies, University of Toronto, North York, ON, Canada, where she leads the Dynamic Systems Laboratory. Her general areas of interest lie at the interface of robotics, controls, and machine learning.
\end{IEEEbiography}

\vspace{-5mm}

\begin{IEEEbiography}[{\includegraphics[width=1in,height=1.25in,clip,keepaspectratio]{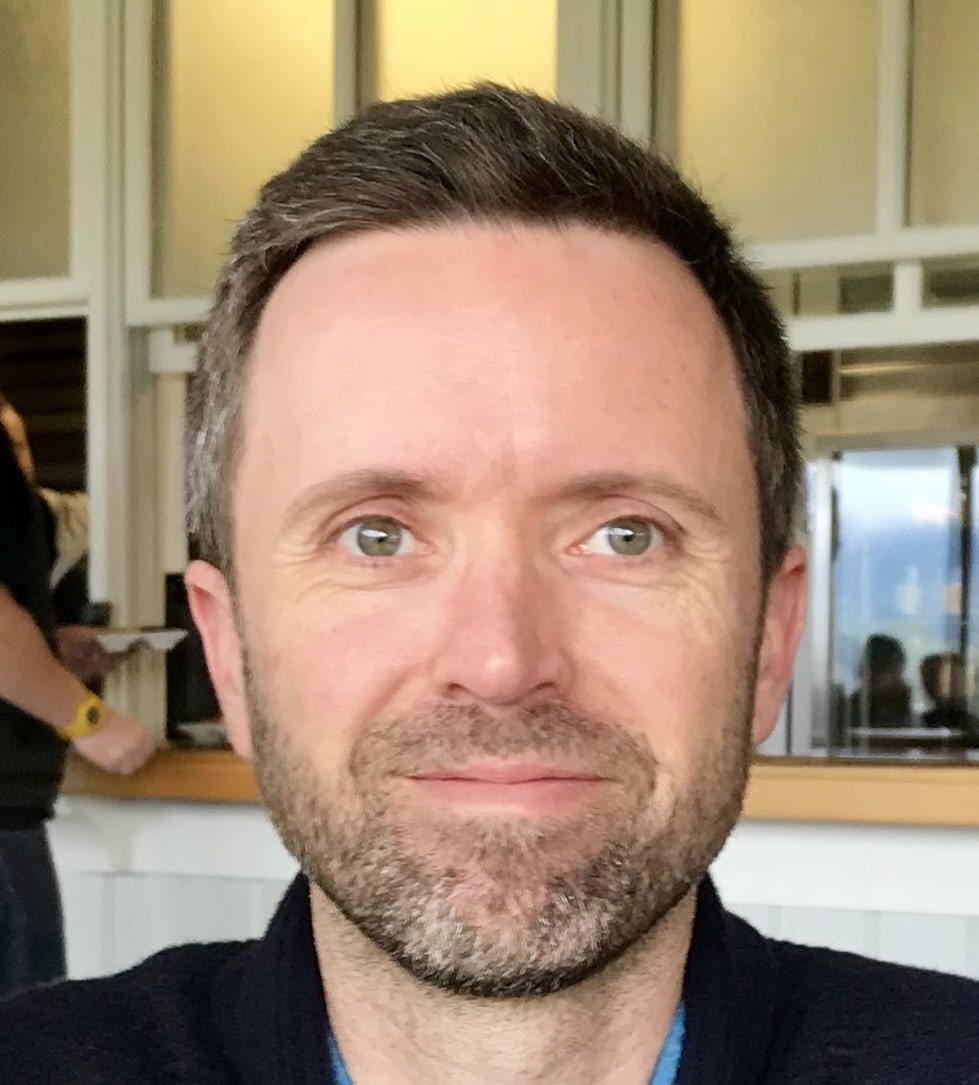}}]{Timothy D. Barfoot} received the B.A.Sc. degree in engineering science from University of Toronto, Toronto, ON, Canada, in 1997 and the Ph.D. degree in aerospace science and engineering from University of Toronto, in 2002. He is a Professor with the University of Toronto Robotics Institute, Toronto, ON, Canada. He works in the areas of guidance, navigation, and control of autonomous systems in a variety of applications. He is interested in developing methods to allow robotic systems to operate over long periods of time in large-scale, unstructured, three-dimensional environments, using rich onboard sensing (e.g., cameras, laser rangefinders, and radar) and computation.
\end{IEEEbiography}\enlargethispage{-10cm}

\end{document}